\PassOptionsToPackage{unicode}{hyperref}
\PassOptionsToPackage{hyphens}{url}
\documentclass[
  11pt,
  letterpaper,
]{article}
\usepackage{lmodern}
\usepackage{amssymb,amsmath}
\usepackage{ifxetex,ifluatex}
\ifnum 0\ifxetex 1\fi\ifluatex 1\fi=0 % if pdftex
  \usepackage[T1]{fontenc}
  \usepackage[utf8]{inputenc}
  \usepackage{textcomp} % provide euro and other symbols
\else % if luatex or xetex
  \usepackage{unicode-math}
  \defaultfontfeatures{Scale=MatchLowercase}
  \defaultfontfeatures[\rmfamily]{Ligatures=TeX,Scale=1}
\fi
\IfFileExists{upquote.sty}{\usepackage{upquote}}{}
\IfFileExists{microtype.sty}{% use microtype if available
  \usepackage[]{microtype}
  \UseMicrotypeSet[protrusion]{basicmath} % disable protrusion for tt fonts
}{}
\makeatletter
\@ifundefined{KOMAClassName}{% if non-KOMA class
  \IfFileExists{parskip.sty}{%
    \usepackage{parskip}
  }{% else
    \setlength{\parindent}{0pt}
    \setlength{\parskip}{6pt plus 2pt minus 1pt}}
}{% if KOMA class
  \KOMAoptions{parskip=half}}
\makeatother
\usepackage{xcolor}
\IfFileExists{xurl.sty}{\usepackage{xurl}}{} % add URL line breaks if available
\IfFileExists{bookmark.sty}{\usepackage{bookmark}}{\usepackage{hyperref}}
\hypersetup{
  hidelinks,
  pdfcreator={LaTeX via pandoc}}
\usepackage[margin=1in,headheight=14pt]{geometry}
\usepackage{color}
\usepackage{fancyvrb}

\DefineVerbatimEnvironment{Highlighting}{Verbatim}{commandchars=\\\{\}}
\usepackage{framed}
\definecolor{shadecolor}{RGB}{248,248,248}
\newenvironment{Shaded}{\begin{snugshade}}{\end{snugshade}}

\newcommand{\ControlFlowTok}[1]{\textcolor[rgb]{0.13,0.29,0.53}{\textbf{#1}}}
\newcommand{\DataTypeTok}[1]{\textcolor[rgb]{0.13,0.29,0.53}{#1}}
\newcommand{\DecValTok}[1]{\textcolor[rgb]{0.00,0.00,0.81}{#1}}

\newcommand{\FunctionTok}[1]{\textcolor[rgb]{0.00,0.00,0.00}{#1}}

\newcommand{\KeywordTok}[1]{\textcolor[rgb]{0.13,0.29,0.53}{\textbf{#1}}}
\newcommand{\NormalTok}[1]{#1}
\newcommand{\OperatorTok}[1]{\textcolor[rgb]{0.81,0.36,0.00}{\textbf{#1}}}

\newcommand{\StringTok}[1]{\textcolor[rgb]{0.31,0.60,0.02}{#1}}

\usepackage{longtable,booktabs}
\usepackage{etoolbox}
\makeatletter
\patchcmd\longtable{\par}{\if@noskipsec\mbox{}\fi\par}{}{}
\makeatother
\IfFileExists{footnotehyper.sty}{\usepackage{footnotehyper}}{\usepackage{footnote}}
\makesavenoteenv{longtable}
\usepackage{graphicx}
\makeatletter
\def\maxwidth{\ifdim\Gin@nat@width>\linewidth\linewidth\else\Gin@nat@width\fi}
\def\maxheight{\ifdim\Gin@nat@height>\textheight\textheight\else\Gin@nat@height\fi}
\makeatother
\setkeys{Gin}{width=\maxwidth,height=\maxheight,keepaspectratio}
\makeatletter
\def\fps@figure{htbp}
\makeatother
\providecommand{\tightlist}{%
  \setlength{\itemsep}{0pt}\setlength{\parskip}{0pt}}
\usepackage{fontspec}
\usepackage{fvextra}
\fvset{breaklines=true,breakanywhere=true,breakindent=1em,fontsize=\scriptsize}

\usepackage{setspace}
\usepackage{xcolor}
\definecolor{academicblue}{HTML}{1F3A5F}
\definecolor{linkblue}{HTML}{2453A6}

\usepackage{titlesec}
\titleformat{\section}{\Large\bfseries\sffamily\color{academicblue}}{\thesection}{0.6em}{}
\titleformat{\subsection}{\large\bfseries\sffamily\color{academicblue!85}}{\thesubsection}{0.5em}{}
\titleformat{\subsubsection}{\normalsize\bfseries\sffamily\color{academicblue!75}}{\thesubsubsection}{0.5em}{}
\titlespacing{\section}{0pt}{20pt}{8pt}
\titlespacing{\subsection}{0pt}{14pt}{6pt}

\usepackage[font=small,labelfont=bf,labelsep=period,textfont=it,justification=justified,singlelinecheck=false]{caption}
\usepackage{booktabs}

\usepackage{float}
\makeatletter
\renewcommand{\fps@figure}{htbp}
\renewcommand{\fps@table}{htbp}
\makeatother
\AtBeginDocument{\hypersetup{colorlinks=true,linkcolor=linkblue,citecolor=linkblue,urlcolor=linkblue,pdfborder={0 0 0}}}

\usepackage{fancyhdr}

\setkeys{Gin}{width=0.80\linewidth,keepaspectratio}

\author{}
\date{}

\begin{document}

\thispagestyle{empty}

\noindent

\colorbox{academicblue}{%
\begin{minipage}[c][45pt][c]{\linewidth}
\color{white}
\hspace{1em}\sffamily\large\bfseries Methodology Paper
\hfill\sffamily\normalsize Tropicals Trait Pipeline
\hfill\sffamily\normalsize 2026-05-30 \,\textbullet\, CC BY 4.0\hspace{1em}
\end{minipage}}

\vspace{2cm}

\begin{center}
{\fontsize{20}{24}\bfseries\sffamily\color{academicblue}\selectfont A Registry-Bound LLM Pipeline for\\[6pt]Evidence-Grounded Trait Extraction across\\[6pt]Tropical Plants, Aquatic Species, and Exotic Pets}

\vspace{1em}
{\fontsize{13}{17}\sffamily\color{academicblue!85}\selectfont 5.5 Million Trait Records from the Tropical Species Encyclopedia}
\end{center}

\vspace{2.5cm}

\begin{center}
{\Large\bfseries Jeff Wang}

\vspace{0.5em}
{\normalsize NEXLY LLC, United States}

\vspace{0.8em}
{\small Correspondence: \href{mailto:jeff@tropicals.cn}{jeff@tropicals.cn}\quad ORCID: \href{https://orcid.org/0009-0001-2905-8439}{0009-0001-2905-8439}}
\end{center}

\vspace{2.5cm}

\begin{center}
\begin{minipage}{0.78\linewidth}
\centering
\small\textit{Builds on the Tropical Species Encyclopedia substrate~[1] and Data Descriptor~[2].}
\end{minipage}
\end{center}

\vfill

\newpage

\hypertarget{abstract}{%
\section{Abstract}\label{abstract}}

We describe a registry-bound large-language-model extraction pipeline
producing evidence-grounded structured trait records at scale, on
cultivated tropical plant, aquatic, and pet species. Four mechanisms
render LLM-derived rows auditable: a versioned 39-key closed-vocabulary
trait registry constraining every admitted value to a typed schema; a
per-row verbatim evidence quote tying each value to source text; a
per-row confidence label (high or medium; low dropped pre-persist); and
multi-version preservation. Applied to 409,880 publishable species from
the Tropical Species Encyclopedia, the pipeline executed 706,220 runs
and persisted 5,489,881 trait records across 409,820 species (99.985\%),
81.57\% at high confidence. We report three validation layers in
descending evidentiary strength: at full population, 90.12\% of
5,427,588 evidence-bearing rows have their quote as a verbatim source
substring (93.49\% excluding one compliance meta-trait); a
quote-supports-value audit on n=100 stratified non-red-zone rows yielded
100/100 (lower bound 96.30\%); face-validity on n=50 red-zone rows
yielded 50/50 Accept (lower bound 92.86\%). Per-record correctness is
not claimed; 100\% pending human curation. The contribution is the
four-mechanism framework.

\textbf{Keywords:} trait extraction; large language models; biodiversity
informatics; cultivated tropical species; Darwin Core; evidence
grounding; registry-bound; methodology paper.

\hypertarget{background-and-summary}{%
\section{1 Background and Summary}\label{background-and-summary}}

\hypertarget{why-structured-trait-data}{%
\subsection{1.1 Why structured trait
data}\label{why-structured-trait-data}}

Structured species traits --- light tolerance, mature size, native
climate, toxicity, propagation method --- are the unit of computation
for a wide range of downstream applications: species filtering and
search, ecological niche modeling, horticultural recommendation engines,
automated identification heuristics, and regulatory compliance routing
(ASPCA {[}9{]} toxicity lists, CITES Appendix membership via Species+
{[}10{]}, IUCN Red List {[}11{]} status). Existing curated trait
databases such as TRY {[}3{]} (plant functional traits), BIEN {[}4{]}
(botanical inventories), GIFT {[}5{]} (global island plant traits), and
AusTraits {[}6{]} have established both the scientific value of
structured trait records and the substantial human cost of producing
them; coverage in those resources is concentrated in plants, drawn
predominantly from manually curated primary literature, and typically
reaches sub-million row scale. None was designed for the cross-domain
commercial-horticulture substrate addressed here.

The Tropical Species Encyclopedia (Wang {[}1{]}; companion Data
Descriptor Wang {[}2{]}) occupies a different niche. It is a
cultivated-taxa substrate spanning tropical plants, freshwater and
marine aquatic species, and exotic pet species, oriented toward both
ecological reference and horticultural / commercial trade. The substrate
provides, for each of \textasciitilde410,000 publishable species, a
free-text Chinese bio organised into seven canonical sections
(morphology, distribution, varieties, habitat, propagation, commercial
value, pests). This text is the raw material from which trait records
are extracted; the substrate itself does not directly publish structured
trait values --- those are the contribution of the pipeline described
here.

\hypertarget{why-llm-trait-extraction-is-a-methodology-problem}{%
\subsection{1.2 Why LLM trait extraction is a methodology
problem}\label{why-llm-trait-extraction-is-a-methodology-problem}}

Producing structured trait data from a 410K-species free-text corpus by
manual curation is infeasible. Large-language-model extraction makes the
scale tractable but introduces a different problem: per-row trust.
Standard LLM failure modes (evidence-unsupported candidate values,
registry-OOV candidates, drift across model versions, miscalibrated
confidence) all manifest at this scale, and an unmediated ``the model
said so'' deposit cannot be safely consumed downstream.

Two further considerations sharpen the methodological framing. First,
the substrate text itself was produced by an upstream LLM (a Qwen-family
model, documented in the companion P1 paper). The extraction pipeline
described here is therefore the \textbf{second hop} in a two-stage LLM
workflow: hop 1 generates the prose, hop 2 extracts typed values from
the prose. P1 addresses hop 1; this paper addresses hop 2. Validation
partitions accordingly: the substring-verification audit reported in §4
grounds hop 2 (does the extracted value's evidence quote actually appear
in the substrate text?) but does not speak to hop 1 (is the substrate
text itself faithful to its upstream sources?). Both hops matter for
end-user trust; making the partition explicit keeps the two questions
separable.

Second, the substrate covers safety-critical categories --- plant and
animal toxicity, CITES Appendix membership, physical hazards. These
cannot be published as undifferentiated LLM outputs even with
provenance. The pipeline therefore routes a designated red-zone subset
of trait keys to a dedicated review queue, itself published as a
transparency artifact of the deposit, and the public consumer surface
gates these traits as informational signals rather than authoritative
determinations (§2.6).

\hypertarget{contribution-and-scope}{%
\subsection{1.3 Contribution and scope}\label{contribution-and-scope}}

The contribution is a four-mechanism extraction framework --- a
closed-vocabulary versioned registry, per-row verbatim evidence quotes,
per-row model-assigned confidence labels, and
\texttt{(species\_id,\ trait\_key,\ model\_version)} UNIQUE preservation
of all model versions --- together with production-scale evidence that
the framework is operable and auditable at 5.5M rows. The paper
documents the framework (§2), the deposit produced under it (§3), and
the validation evidence available at 5.5M scale without per-record human
review (§4).

What this paper deliberately does not do: it does not claim per-record
correctness for any individual trait row; it does not report Cohen's κ
inter-rater agreement (a stratified audit sample is published with the
deposit to enable external blind review, and the principal validation is
identified as deferred); and it does not address hop-1 fidelity, which
is the substrate paper's scope (Wang {[}1, 2{]}). The deposit is
positioned as an audit-ready substrate over which downstream consumers
can apply their own stratification (by confidence, by trait key, by
model\_version, by red-zone status) for the level of trust their use
case requires.

\hypertarget{methods}{%
\section{2 Methods}\label{methods}}

This section describes the protocol by which a closed trait registry, a
registry-bound LLM extractor, and a two-stage server-side admission gate
produced 5,489,881 structured trait assertions over 409,820 cultivated
tropical species (99.985\% of the 409,880-species publishable
substrate). Counts are recomputed from the database snapshot of
2026-05-29; schema DDL, the full 39-key registry, and operational
failure modes are in §S1--§S3.

\hypertarget{substrate-and-scope}{%
\subsection{2.1 Substrate and scope}\label{substrate-and-scope}}

The extraction substrate is the published \emph{Tropicals.cn: Tropical
Species Encyclopedia} dataset (Wang {[}1{]}; companion Data Descriptor
Wang {[}2{]}). It contains 410,499 active species records, of which
409,880 satisfy the publishable gate --- an eleven-clause predicate over
completeness of bilingual nomenclature, taxonomic placement, origin
geography, and seven required \texttt{bio\_sections} keys. This
409,880-species subset is the denominator throughout, partitioned by the
\texttt{category} enum into \texttt{tropical\_plants} (271,786, 66.3\%),
\texttt{tropical\_pets} (89,521, 21.8\%), and \texttt{tropical\_aquatic}
(48,573, 11.9\%).

Each species' input is a single JSON column,
\texttt{species\_detail.bio\_sections}, with seven fixed keys
(\texttt{morphology}, \texttt{distribution}, \texttt{varieties},
\texttt{habitat}, \texttt{propagation}, \texttt{commercial},
\texttt{pests}). Each key holds a Chinese-language section of roughly
150--300 characters; a typical species supplies 1,000--2,500 characters
total. Section keys are stable across subdomains; content semantics
adapt (e.g.~\texttt{habitat} carries water chemistry for aquatic species
and substrate composition for plants).

A consequence of the substrate's design merits framing. The
\texttt{bio\_sections} payload is itself LLM-generated upstream ---
produced by Qwen \texttt{qwen3.6-plus} under structured prompting
against curated source catalogues. \textbf{The present paper's
contribution is therefore the second hop of a two-hop LLM pipeline}:
structured trait extraction from already-deposited substrate text. The
extractor's correctness is bounded above by the substrate's: a trait
inconsistent with primary literature is correctly extracted from a
payload that already encodes that inconsistency. The
substring-verification audit in §4.2 verifies hop-2 grounding only;
hop-1 quality is treated as fixed input. Cultivar, hybrid, and seedling
identifiers (775 + 371 records in upstream staging) are deferred to a
schema-version increment; the deposit is species-level only.

\hypertarget{trait-registry}{%
\subsection{2.2 Trait registry}\label{trait-registry}}

Traits are organised under a versioned registry
(\texttt{scripts/mimo/trait-registry.mjs}) enumerating 39 trait keys
across four applicability domains and five value types. The full per-key
specification --- allowed-value lists, integer/range bounds, evidence
hints --- is in §S1.

\textbf{Table 1.} Registry structure: trait-key counts by applicability
domain × value type.

\begin{longtable}[]{@{}lrrrrrr@{}}
\toprule
domain & text & enum & multi\_enum & int & range & total\tabularnewline
\midrule
\endhead
all (universal) & 4 & 8 & 6 & 0 & 0 & 18\tabularnewline
plants & 0 & 3 & 2 & 1 & 1 & 7\tabularnewline
aquatic & 0 & 2 & 1 & 1 & 3 & 7\tabularnewline
pets & 0 & 2 & 1 & 1 & 3 & 7\tabularnewline
\textbf{total} & \textbf{4} & \textbf{15} & \textbf{10} & \textbf{3} &
\textbf{7} & \textbf{39}\tabularnewline
\bottomrule
\end{longtable}

Each trait key declares exactly one value type. \texttt{enum} carries a
closed list of mutually exclusive tokens; \texttt{multi\_enum} a
delimited set from a closed list; \texttt{text} a short free-form string
under a per-key character ceiling, for attributes that cannot be
enumerated in advance (\texttt{native\_region\_primary},
\texttt{mature\_size\_text}); \texttt{int} a bounded integer
(e.g.~\texttt{min\_temperature\_c} in {[}-10, 30{]} °C); \texttt{range}
a string-encoded \texttt{"x–y"} interval. The single-value-type-per-key
contract is invariant.

Four registry keys carry a \texttt{red\_zone=true} flag, encoding traits
whose incorrect values carry materially heavier downstream consequences:
\texttt{toxicity\_to\_humans}, \texttt{toxicity\_to\_pets},
\texttt{physical\_hazards}, \texttt{cites\_appendix\_in\_bio}. A
canonical safety list outside the registry names a fifth,
\texttt{iucn\_status}, as red-zone, but it is not among the 39
registered targets and has zero persisted rows --- the gap is
intentional for v1 (the upstream substrate does not consistently surface
IUCN listings) and is a scope-expansion item.

The registry declares
\texttt{schema\_version\ =\ \textquotesingle{}v1\textquotesingle{}}; all
5,489,881 rows carry this tag. A future \texttt{v2} would introduce new
rows under a distinct label without rewriting v1 rows. The JSON snapshot
shipped with the deposit is named \texttt{trait\_registry\_v2.json}; the
\texttt{\_v2} suffix denotes the JSON snapshot's structural format
version (key order, allowed-value vocabulary layout), distinct from the
dataset's \texttt{schema\_version} field which remains \texttt{v1} in
this release.

\hypertarget{extraction-pipeline}{%
\subsection{2.3 Extraction pipeline}\label{extraction-pipeline}}

The extractor is Xiaomi MiMo \texttt{mimo-v2.5} via an OpenAI-compatible
chat-completions API across two regional endpoints
(\texttt{token-plan-cn}, \texttt{token-plan-sgp}). The canonical
production tag
\texttt{model\_version=\textquotesingle{}full-v1-20260524\textquotesingle{}}
accounts for 5,487,270 (99.95\%) of deposited rows; three smaller tags
(2,431 + 157 + 23) coexist under the unique-key contract in §2.5.

The base \texttt{mimo-v2.5} was selected over \texttt{mimo-v2.5-pro}
after a controlled head-to-head on 19 species under identical prompts:
the base produced a higher fraction of high-confidence rows (84.9\%
vs.~80--82\%) and sustained roughly twice the throughput at cluster
width (3.0--4.0 species/s vs.~1.5--2.0), with per-row quality
statistically indistinguishable; \texttt{pro} additionally exhibited a
three-hour HTTP 502 episode under sustained load. Registry-bound
structured-JSON extraction over short Chinese source text is a setting
in which \texttt{pro}'s reasoning-token allocation is not productively
spent. The cluster runs eight processes with total concurrency capped at
280 outstanding calls and per-key burst at 40--50; a four-state key pool
(\texttt{active}, \texttt{standby}, \texttt{cooldown}, \texttt{dead})
absorbs per-key platform variability (§S3).

Each call passes one species' \texttt{bio\_sections} block with the
registry subset applicable to its subdomain (18 universal keys + 7 for
the row's \texttt{category} = 25 candidates per call). The trait-key
vocabulary is named in the prompt and allowed-value lists summarised,
but the binding contract is enforced by the post-processing layer in
§2.4, not by prompt instruction. The model returns a JSON array of
\texttt{(trait\_key,\ value,\ evidence\_quote,\ evidence\_section,\ ai\_reasoning,\ confidence)}
tuples. The prompt routes low-confidence candidates to null returns, so
no \texttt{low}-confidence rows enter the deposit by construction; the
model may abstain on any candidate, with abstentions counted per run.

\begin{figure}
\centering
\includegraphics{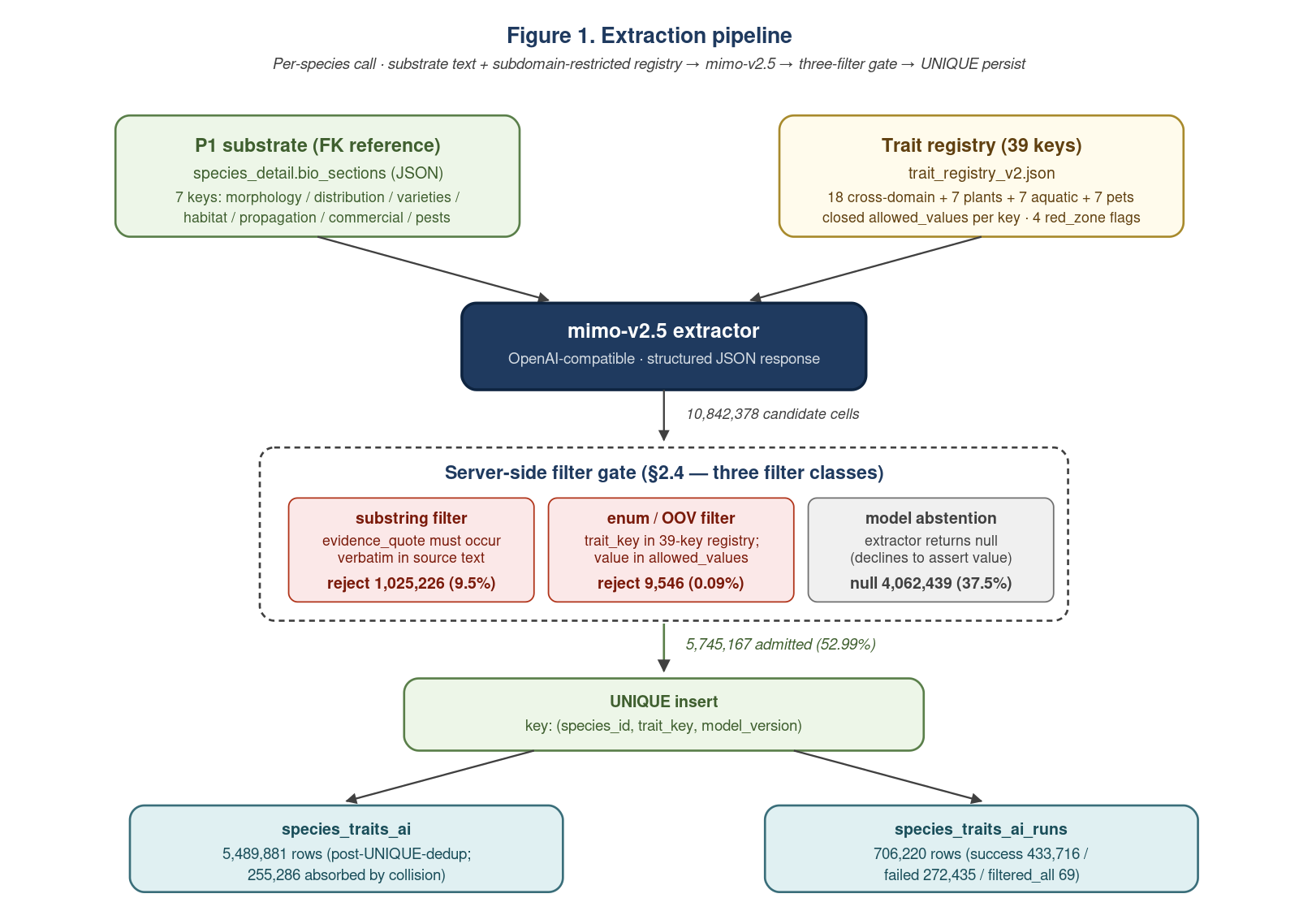}
\caption{The extraction pipeline. Each species-level substrate record is
passed with the subdomain-restricted 39-key registry to the
\texttt{mimo-v2.5} extractor; the structured response is admitted only
after passing the substring-verification and enum-conformance filters
before persistence. Per-run telemetry is written separately to
\texttt{species\_traits\_ai\_runs}.}
\end{figure}

\hypertarget{filtering-and-admission-the-three-filter-classes}{%
\subsection{2.4 Filtering and admission --- the three filter
classes}\label{filtering-and-admission-the-three-filter-classes}}

Two server-side filter stages separate the model's response from row
insertion; a third non-admission path --- model abstention --- accounts
for the largest share of unproduced cells. Reporting these three classes
separately, not as a single conflated ``reject'' bucket, is the headline
accounting commitment of this paper.

The \textbf{substring-verification filter} (anti-fabrication of
evidence) requires that the emitted \texttt{evidence\_quote} occur as an
exact substring of the source \texttt{bio\_sections} payload. Failed
candidates are counted in \texttt{traits\_filtered\_hallucination}
(legacy DDL column name; the paper refers to this as the substring /
evidence-unsupported filter).

The \textbf{registry-OOV / enum-conformance filter} rejects two
out-of-vocabulary patterns: an emitted \texttt{trait\_key} not in the
39-key registry, or for \texttt{enum}/\texttt{multi\_enum} keys an
emitted value not in the allowed-value list at the run's
\texttt{schema\_version} (for \texttt{int}/\texttt{range} keys, the
value must lie within declared bounds). Failed candidates are counted in
\texttt{traits\_filtered\_enum} (legacy DDL column name; there is no
separate \texttt{traits\_filtered\_oov} column --- the enum filter
\emph{is} the OOV path, covering both key-level and value-level
violations).

The third path is \textbf{model abstention}: the extractor returns no
value when the source supports no assertion at any confidence.
Abstentions are counted in \texttt{traits\_returned\_null} and disclosed
as the dominant non-admission path, not folded into a reject total.
Abstention is a protocol feature --- the model is permitted to leave a
value blank rather than fabricate --- and reporting its share separately
is necessary for any consumer reasoning about per-species coverage.

\textbf{Table 2.} Candidate-cell outcome decomposition across the
433,716 successful runs.

\begin{longtable}[]{@{}lrr@{}}
\toprule
Outcome & Count & Share\tabularnewline
\midrule
\endhead
Admitted (passed both filters) & \textbf{5,745,167} &
\textbf{52.99\%}\tabularnewline
Substring / evidence-unsupported reject & 1,025,226 &
9.46\%\tabularnewline
Registry-OOV / enum reject & 9,546 & 0.09\%\tabularnewline
Model abstention & 4,062,439 & 37.47\%\tabularnewline
\textbf{Total candidates emitted} & \textbf{10,842,378} &
\textbf{100.00\%}\tabularnewline
\bottomrule
\end{longtable}

Two findings merit notice. First, the two rejection mechanisms are
dramatically asymmetric: substring grounding rejects two orders of
magnitude more candidates than enum-conformance, consistent with tight
allowed-value lists strongly constraining generation while leaving more
room for under-supported assertions. Second, abstention dominates
non-admission at 37.47\% --- comparable in magnitude to admission itself
--- and is correctly understood as the extractor declining to assert
under-supported claims, not a failure; conflating it with rejection
would inflate the apparent reject rate fourfold.

Of the 5,745,167 admitted cells, \textbf{5,489,881 are persisted as
distinct rows}; the 255,286 difference (4.4\%) is absorbed by the
\texttt{(species\_id,\ trait\_key,\ model\_version)} UNIQUE constraint.
Across the 433,716 successful runs over 409,820 species (≈24K species
received more than one successful run --- typically retry sessions or
pipeline-validation re-extractions), repeat runs re-emit the same
\texttt{(species\_id,\ trait\_key,\ model\_version)} cells; the UNIQUE
constraint absorbs the second and subsequent emissions, accounting for
the 255,286 admitted-but-not-persisted rows. Filter counters are written
per run to \texttt{species\_traits\_ai\_runs}, not per row --- a
consumer auditing schema conformance reads the trait table (every row
passed both filters by construction); a consumer studying model
behaviour reads the run table. Neither filter is a per-row correctness
audit; §4.2 only verifies that the evidence quote appears verbatim in
substrate text, not that it supports the asserted value, nor that the
value itself is correct.

\hypertarget{persistence-and-provenance}{%
\subsection{2.5 Persistence and
provenance}\label{persistence-and-provenance}}

Each surviving cell is inserted into \texttt{species\_traits\_ai} with
seven provenance columns alongside the core
\texttt{(species\_id,\ trait\_key,\ value,\ value\_type)} tuple ---
model and schema version tags, evidence quote and section pointer, model
reasoning, confidence label, and extraction timestamp (full schema in
§S2). The primary unique key is
\texttt{(species\_id,\ trait\_key,\ model\_version)} --- chosen so that
re-extraction under a new model tag appends rather than replaces,
preserving cross-model history. Of 5,485,630 distinct
\texttt{(species\_id,\ trait\_key)} pairs at exactly one model\_version,
1,941 carry two and 123 carry three; among the 2,064 multi-version
pairs, 1,518 (73.5\%) produced identical values and 546 (26.5\%)
diverged, with divergent values inspectable under their respective tags.

Evidence coverage is high: 98.87\% carry a non-empty
\texttt{evidence\_quote}, 98.97\% \texttt{evidence\_section}, ≈100\%
\texttt{ai\_reasoning}. The \texttt{evidence\_quote} distribution is
concentrated at short Chinese phrases --- 98.2\% are 1--50 characters,
mean length 17 --- consistent with the extractor pointing at a specific
phrase, the pattern that makes the substring audit tractable at scale.
The \texttt{evidence\_section} field is \texttt{varchar(32)} free text;
eight canonical values (\texttt{habitat}, \texttt{morphology},
\texttt{distribution}, \texttt{commercial}, \texttt{propagation},
\texttt{quick\_card}, \texttt{varieties}, \texttt{pests}) plus empty
account for 99.99\%, with a long tail of \textasciitilde60 typographic
variants at \textless0.02\% preserved verbatim.

Confidence is a two-level enum: \textbf{4,477,850 rows (81.57\%) at
\texttt{high}, 1,012,031 (18.43\%) at \texttt{medium}}; no row carries
\texttt{low} (routed to null returns). High-confidence concentration
varies by trait key: among five quantitative keys reading a directly
stated numerical value (e.g.~\texttt{optimal\_temp\_range\_c},
\texttt{enclosure\_humidity\_pct}, \texttt{water\_ph\_range}) the rate
exceeds 99\%; among inference-heavy keys (\texttt{growth\_rate\_tier},
\texttt{popularity\_signal}, \texttt{propagation\_difficulty}) it falls
to 41--67\%.

Aggregate run telemetry across the working window (2026-05-24 to
2026-05-29 UTC) records \textbf{706,220 calls} under a mutually
exclusive \texttt{species\_traits\_ai\_runs.status} enum
(\texttt{success} / \texttt{failed} / \texttt{filtered\_all}; a
\texttt{timeout} value declared in the DDL is never observed). Run-level
outcomes: \textbf{433,716 runs successfully wrote at least one admitted
trait row; 272,435 runs failed} (no candidate cells emitted, typically
from transient infrastructure errors); \textbf{69 runs completed with
\texttt{filtered\_all} status} (all emitted cells were rejected by the
substring or enum filter and none persisted). Sum: 706,220. The 38.58\%
failed-run rate is expected operational behaviour for a hosted-LLM
fan-out cluster --- transient HTTP 429/502 absorbed by the key pool ---
and does not reflect extraction quality, because the resume protocol
re-attempts any species lacking a \texttt{success} or
\texttt{filtered\_all} row at the current model version. The 60 species
not covered (409,820/409,880, \textbf{99.985\%}) are those for which no
run ever terminated in \texttt{success} over the working window.
Successful runs averaged 91 s and aggregated 4.93 B tokens (1.33 B
\texttt{input\_tokens}, 2.26 B \texttt{completion\_tokens}, 1.34 B
\texttt{reasoning\_tokens}).

\hypertarget{red-zone-routing-and-ethics-of-public-unverified-distribution}{%
\subsection{2.6 Red-zone routing and ethics of public-unverified
distribution}\label{red-zone-routing-and-ethics-of-public-unverified-distribution}}

Every persisted row carries an \texttt{admin\_review\_status} field
(enum over \texttt{pending}, \texttt{approved}, \texttt{rejected},
\texttt{flagged}) indexed by
\texttt{(admin\_review\_status,\ trait\_key)} so that a curator can
scope a working set to one trait-key's pending rows with a single
indexed predicate. At the snapshot date all 5,489,881 rows are
\texttt{pending}; the extraction sweep landed rows under the default
before any moderator sweep began. The deposit documents the queue's
design and its contents at extraction time, not outcomes of review.

Red-zone routing is a per-trait-key flag on the registry, not a separate
persistence path. The four red-zone keys ---
\texttt{cites\_appendix\_in\_bio} (303,786), \texttt{physical\_hazards}
(105,757), \texttt{toxicity\_to\_humans} (12,435),
\texttt{toxicity\_to\_pets} (6,690), totalling \textbf{428,668 rows} ---
are persisted into \texttt{species\_traits\_ai} with the same evidence
and confidence treatment as non-red-zone rows; the moderation-queue
index orders them ahead of other trait keys in the curator UI. The
red-zone high-confidence rate is reported separately:
\textbf{376,468/428,668 (87.82\%) at \texttt{high}} --- 6.25 percentage
points above the 81.57\% global rate, not marginally so --- with
substantial variation across keys (98.80\%
\texttt{cites\_appendix\_in\_bio}, 63.60\% \texttt{physical\_hazards},
50.10\% \texttt{toxicity\_to\_humans}, 41.50\%
\texttt{toxicity\_to\_pets}). CITES status is an explicit lexical signal
and extracts at high confidence; pet-toxicity assertions require
inference from morphological and ecological cues and bear a heavier
downstream review obligation.

\begin{figure}
\centering
\includegraphics{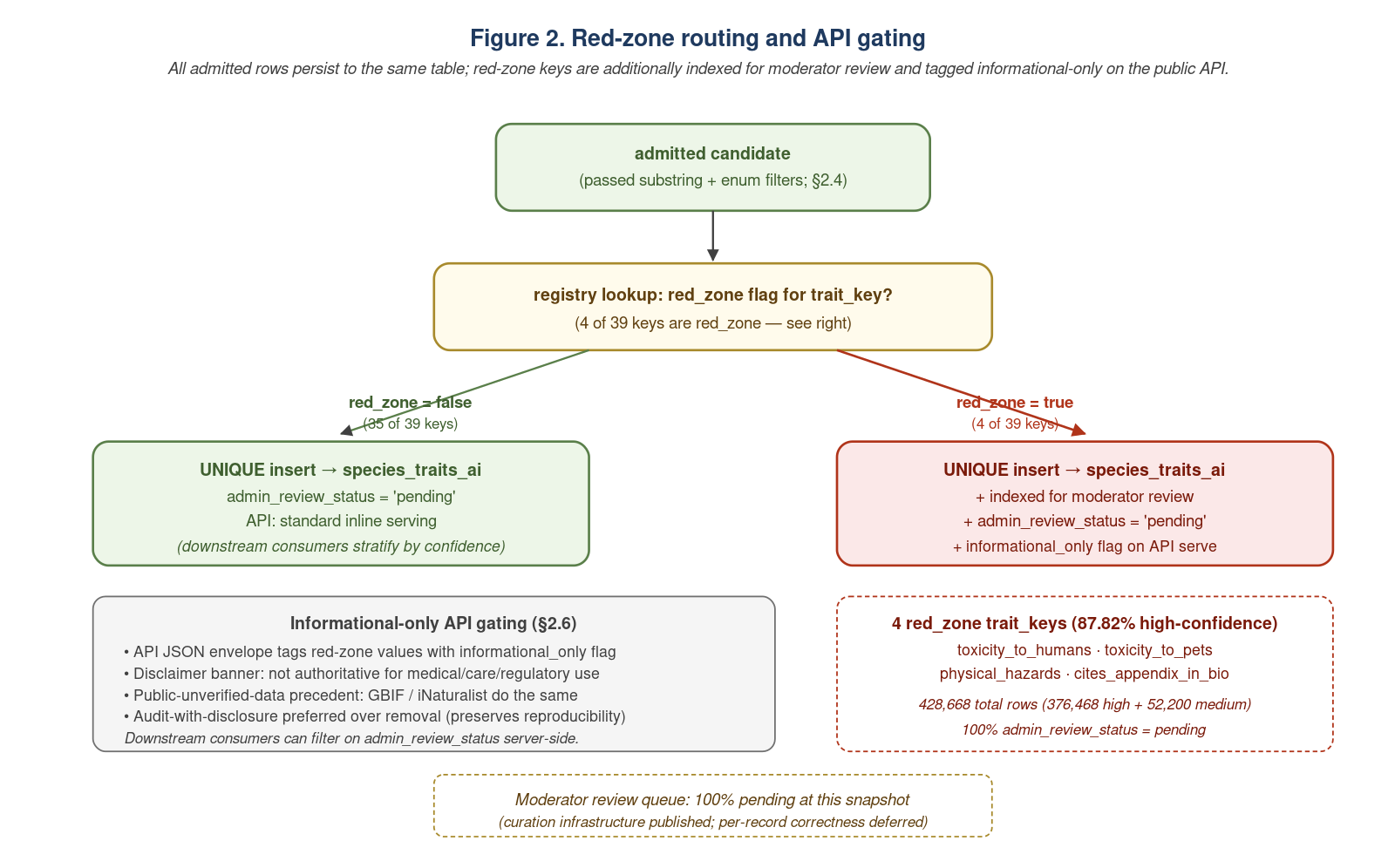}
\caption{Red-zone routing. Registry-flagged red-zone keys (4 of 39) are
persisted into the standard trait table but indexed for priority
moderator review; the index pre-orders curator effort onto
safety-bearing keys without altering the extraction or persistence path.
Red-zone high-confidence rate (87.82\%) exceeds the global rate
(81.57\%) by 6.25 pp.}
\end{figure}

\textbf{Ethics of public distribution under 100\% pending review.}
Red-zone rows are exposed through public \texttt{/api/v1} endpoints
alongside non-red-zone rows while all 428,668 remain \texttt{pending}.
The risk is real and worth naming: an unreviewed assertion that a
species is non-toxic, or that its \texttt{physical\_hazards} are
\texttt{none}, could be misused as care or medical guidance. We
nonetheless publish, with three defences. First, every API response
carries an explicit \texttt{informational\_only} flag and a structured
disclaimer in the JSON envelope, and \texttt{admin\_review\_status} is
itself a response field --- consumers can filter on \texttt{approved}
once the queue accumulates non-pending values. Second, public
distribution of unreviewed data under provenance disclaimers is the
established convention in adjacent biodiversity infrastructure: GBIF
{[}7{]} and iNaturalist {[}8{]} publish unverified taxonomic and
observational records at scale under disclaimer-as-gating, and the
community has converged on this norm for the openness it enables. Third,
the conservative choice in a reproducibility-bearing deposit is
auditability with disclosure rather than silent removal: hiding red-zone
rows would destroy consumers' ability to audit which assertions exist
and on what evidence. We therefore publish with disclosure, queue
visibility, and downstream filterability; alternative gating
(authenticated access, removal pending review, two-tier APIs) is out of
scope at v1 and remains an escalation path if disclaimer-as-gating
proves insufficient.

\hypertarget{data-records}{%
\section{3 Data Records}\label{data-records}}

The trait-extension dataset described here is published as an
independent Zenodo record with its own Concept DOI (issued at first
publication; subsequent releases use the same Concept DOI through
Zenodo's \texttt{newversion} workflow). The substrate is the Tropical
Species Encyclopedia (Wang {[}1{]}; companion Data Descriptor Wang
{[}2{]}); the trait extension is never published as a new version of the
substrate record, so the two deposits are versioned independently. The
snapshot characterised below is dated 2026-05-29 against the working
population \texttt{publishable\ =\ 1}, comprising 409,880 species.
Original-contribution layers are released under CC-BY 4.0;
identifier-only references to the substrate preserve that record's
licence discipline, and no upstream descriptive text is duplicated here.

\hypertarget{database-tables}{%
\subsection{3.1 Database tables}\label{database-tables}}

The deposit comprises two production tables plus a JSON snapshot of the
trait registry. Figure 3 shows how they attach to the P1 substrate via
the shared identifier \texttt{species\_id}.

\begin{figure}
\centering
\includegraphics{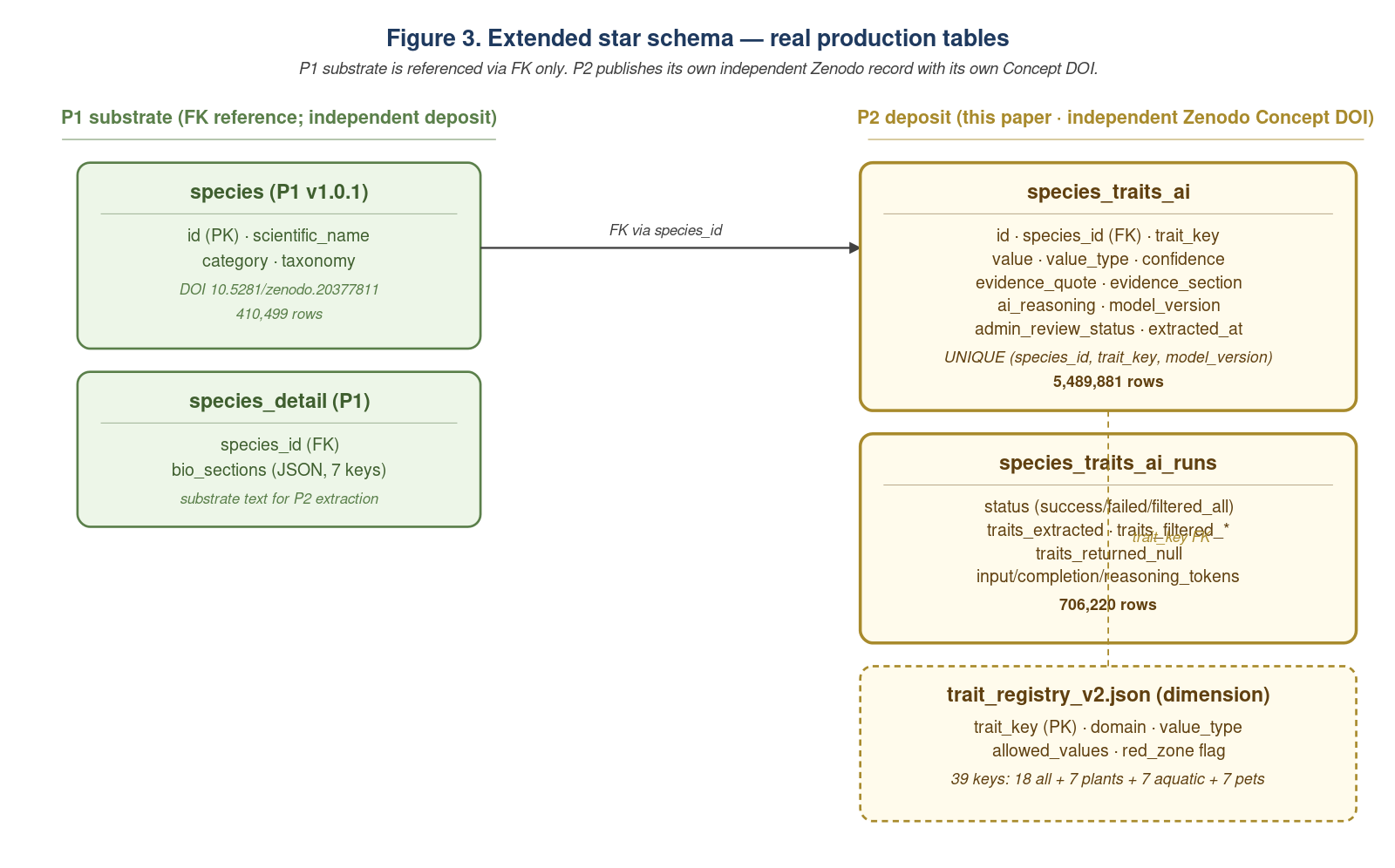}
\caption{Extended star schema --- \texttt{core\_taxon} (P1 substrate,
referenced via FK) and the two P2 trait tables
\texttt{species\_traits\_ai} and \texttt{species\_traits\_ai\_runs}. P2
publishes its own independent Zenodo record; substrate references are
id-only.}
\end{figure}

\textbf{\texttt{species\_traits\_ai}} --- the trait fact table. One row
per \texttt{(species\_id,\ trait\_key,\ model\_version)} triple admitted
under the registry gate and substring grounding filter described in §2.
Columns: \texttt{id} (BIGINT, PK), \texttt{species\_id} (BIGINT, FK →
\texttt{species(id)} via \texttt{ON\ DELETE\ CASCADE}),
\texttt{trait\_key} (VARCHAR(64), one of 39 registry keys),
\texttt{value} (VARCHAR(255), stored verbatim with type-specific parsing
performed downstream via \texttt{value\_type}), \texttt{value\_type}
(ENUM: \texttt{enum}, \texttt{multi\_enum}, \texttt{int},
\texttt{range}, \texttt{text}, \texttt{bool} --- \texttt{bool} reserved
but unused), \texttt{confidence} (ENUM: \texttt{high}, \texttt{medium}
--- \texttt{low}-confidence candidates are filtered at extraction time
and not persisted), \texttt{evidence\_quote} (VARCHAR(500), a verbatim
substring of the source paragraph), \texttt{evidence\_section}
(VARCHAR(32), source-paragraph key, normally one of the seven canonical
\texttt{bio\_sections} keys or the \texttt{quick\_card} pseudo-section),
\texttt{ai\_reasoning} (VARCHAR(300)), \texttt{model\_version}
(VARCHAR(64)), \texttt{schema\_version} (VARCHAR(16)),
\texttt{admin\_review\_status} (ENUM: \texttt{pending},
\texttt{approved}, \texttt{rejected}, \texttt{flagged}), and
\texttt{extracted\_at} (DATETIME). The table is governed by
\texttt{UNIQUE\ KEY\ (species\_id,\ trait\_key,\ model\_version)}, so
re-extraction under a new checkpoint preserves the prior extraction
rather than overwriting it --- multi-version histories are first-class.
The snapshot contains 5,489,881 rows across 409,820 species; integrity
verified with zero orphan rows, zero unique-key duplicates, and zero
NULLs in required columns.

\textbf{\texttt{species\_traits\_ai\_runs}} --- the per-extraction-run
telemetry log. One row per attempted extraction
\texttt{(species\_id,\ model\_version)} regardless of outcome, with a
mutually exclusive \texttt{status} enum (\texttt{success},
\texttt{failed}, \texttt{filtered\_all}; \texttt{timeout} declared but
never observed), per-class filter counters (admitted / substring-reject
/ enum-reject / abstention), token usage and timing fields, and the
truncated error message plus the full retained \texttt{raw\_response}
payload for post-hoc gate re-evaluation (full schema in §S2). The
snapshot contains 706,220 runs covering every publishable species at
least once. The legacy column names
\texttt{traits\_filtered\_hallucination} and
\texttt{traits\_filtered\_enum} reflect the pipeline's earlier
development; narrative text uses the operationally accurate terms
``substring filter'' and ``enum/registry-OOV filter''.

\textbf{\texttt{species\_detail.bio\_sections}} --- not deposited with
P2. The substrate paragraphs read by the extractor live on
\texttt{species\_detail.bio\_sections} as a JSON column in the P1
deposit (seven canonical keys: \texttt{morphology},
\texttt{distribution}, \texttt{varieties}, \texttt{habitat},
\texttt{propagation}, \texttt{commercial}, \texttt{pests}). The trait
layer references it by foreign key only; downstream users auditing
individual extractions retrieve the source paragraph from the P1 deposit
using \texttt{species\_id} and the \texttt{evidence\_section} key.

The trait registry itself lives in the platform source code as a
JavaScript module; the deposit ships a JSON snapshot
(\texttt{trait\_registry\_v2.json}) capturing all 39 trait entries
together with the verbatim source module under \texttt{code/} for full
reproducibility of the gate logic.

\hypertarget{distribution-by-domain-and-category}{%
\subsection{3.2 Distribution by domain and
category}\label{distribution-by-domain-and-category}}

The 39 trait keys are partitioned by registry domain
(\texttt{applies\_to}): 18 universal (\texttt{all}), 7
plant-specialised, 7 pet-specialised, and 7 aquatic-specialised. By
extracted volume these break down as 3,266,199 rows for \texttt{all}
traits (59.5\%), 1,609,818 for \texttt{plants} (29.3\%), 375,621 for
\texttt{pets} (6.8\%), and 238,243 for \texttt{aquatic} (4.3\%). The mix
tracks substrate composition --- publishable species are predominantly
\texttt{tropical\_plants} (271,786, 66.3\%), followed by
\texttt{tropical\_pets} (89,521, 21.8\%) and \texttt{tropical\_aquatic}
(48,573, 11.9\%) --- modulated by the additional weight from the 18
universal traits that fire on every species. Within each category,
coverage rounds to 100.0\% at one decimal place; exact corpus-wide
coverage is 99.985\% (409,820 of 409,880 publishable species carry at
least one persisted trait row).

By \texttt{value\_type} the snapshot decomposes into 2,154,628
\texttt{enum} rows (15 keys), 1,765,357 \texttt{multi\_enum} (10 keys),
747,174 \texttt{text} (4 keys), 592,205 \texttt{range} (7 keys), and
230,517 \texttt{int} (3 keys). Every trait key maps to exactly one
\texttt{value\_type}, with no cross-type inconsistency observed.

\hypertarget{model-version-and-confidence-distribution}{%
\subsection{3.3 Model version and confidence
distribution}\label{model-version-and-confidence-distribution}}

The snapshot is dominated by a single canonical extraction batch. The
\texttt{full-v1-20260524} checkpoint accounts for 5,487,270 rows
(99.95\%); a pilot batch \texttt{pilot-v1-20260524-1304} contributes
2,431 rows; a small cross-family fix-up batch using
\texttt{deepseek-v4-flash-fix-20260529} contributes 157 rows; and
\texttt{smoke-test-130109} retains 23 rows preserved in the deposit for
audit trail rather than removed. All 5,489,881 rows carry
\texttt{schema\_version\ =\ \textquotesingle{}v1\textquotesingle{}}.

Model-assigned high-confidence rows account for 4,477,850 (81.57\%) and
medium-confidence rows for the remaining 1,012,031 (18.43\%). This is
the confidence the extractor model itself assigned; candidates with
model-assigned \texttt{low} confidence are filtered at extraction time
and not persisted. The confidence label is a model self-report and is
\emph{not} a substitute for human review. Every row in the snapshot has
\texttt{admin\_review\_status\ =\ \textquotesingle{}pending\textquotesingle{}}:
the four-valued review enum, the admin moderation UI, and the red-zone
prioritisation queue are all provisioned and operational, but no human
curation has been applied at this snapshot. We report this honestly as a
pipeline-stage fact rather than a defect; aggregate quality signals
independent of human review are documented in §4.

\hypertarget{coverage-matrix-and-trait-rows-per-species}{%
\subsection{3.4 Coverage matrix and
trait-rows-per-species}\label{coverage-matrix-and-trait-rows-per-species}}

Table 6 reports per-trait row counts and the model-assigned
high-confidence share (\texttt{pct\_high}) for each of the 39 registered
keys, grouped by trait domain (universal / plants / aquatic / pets).
High-confidence concentration is greatest on universally-firing traits
whose evidence is lexically explicit in the source bio
(\texttt{native\_region\_primary}, \texttt{native\_climate\_type},
\texttt{cites\_appendix\_in\_bio}) and lower on inference-heavy keys
(\texttt{propagation\_difficulty}, \texttt{popularity\_signal},
pet-toxicity assertions). Row counts vary by trait scope: universal keys
fire on every publishable species, while subdomain keys (plant / aquatic
/ pet) are bounded above by their in-scope category populations (271,786
/ 48,573 / 89,521). For example, \texttt{enclosure\_temp\_day\_c}
carries 87,857 rows against an in-scope denominator of 89,521 pet
species (≈98\% within scope); read against the full 409,880-species
substrate it is ≈21\%.

\textbf{Table 6.} Per-trait coverage and model-assigned high-confidence
share, ordered by trait domain (universal / plants / aquatic / pets).
Rendered as a tabular visualization for compactness; underlying numbers
in §S5.

\begin{figure}
\centering
\includegraphics{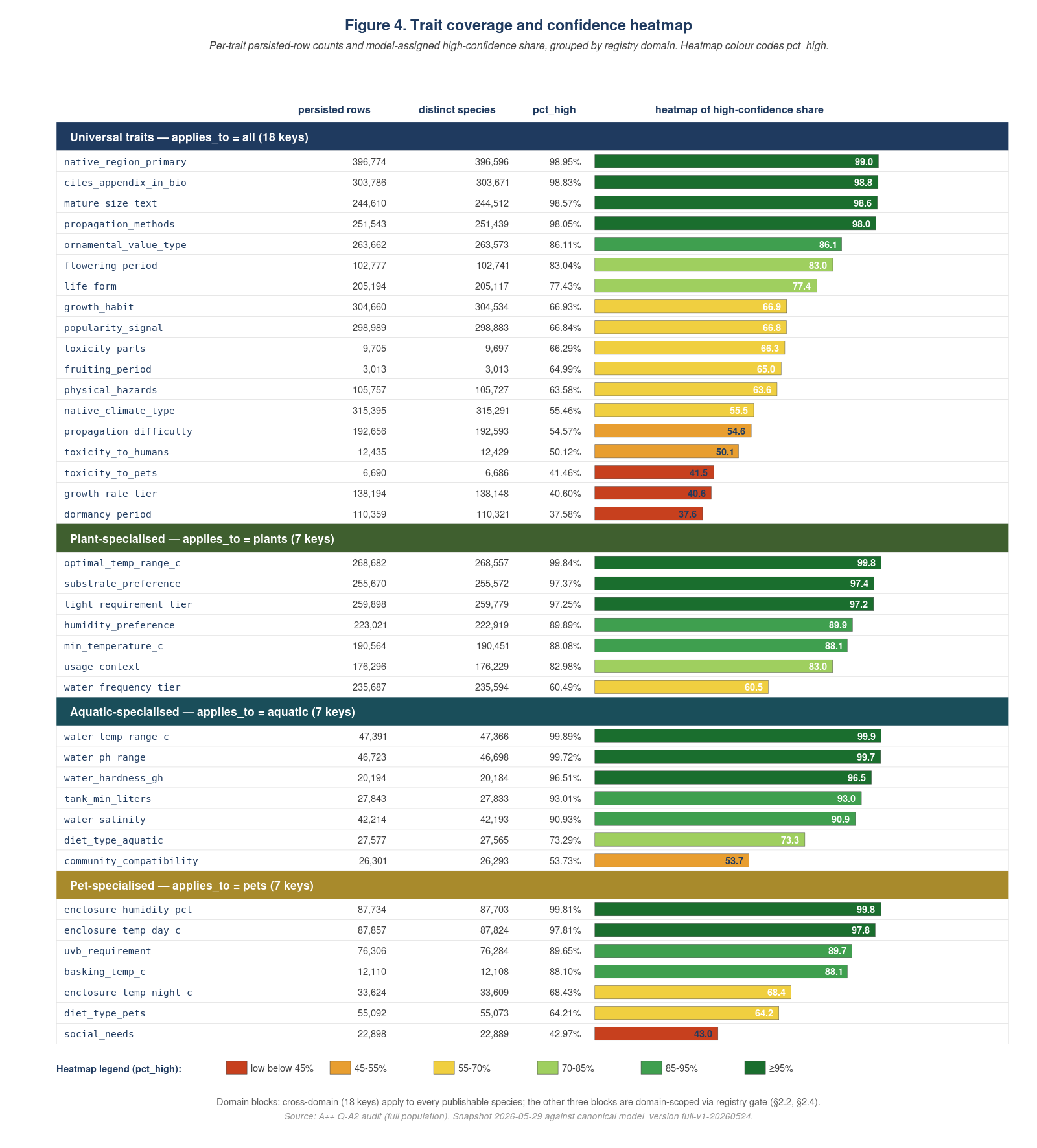}
\caption{Per-trait coverage and high-confidence share by trait\_key,
grouped by trait domain. Snapshot 2026-05-29 against canonical
model\_version \texttt{full-v1-20260524}.}
\end{figure}

The trait-rows-per-species distribution is tight: 280,506 publishable
species carry 11--15 trait rows, 80,125 carry 16--25, 47,952 carry
6--10, 1,120 carry 1--5, 117 carry 26--50, and 60 carry zero. The sum
reconciles to 409,880 publishable species.

\hypertarget{release-artifacts-and-access}{%
\subsection{3.5 Release artifacts and
access}\label{release-artifacts-and-access}}

The deposit is published as a new Zenodo record with its own Concept DOI
(issued at first publication; subsequent versions through the
\texttt{newversion} workflow under the same Concept DOI). The substrate
(P1) is referenced via id-only FK; the trait extension is never
published as a new version of the substrate's Concept DOI (Wang
{[}1{]}), so the two deposits are versioned independently. Snapshot
date: 2026-05-29.

The release ships the following primary artifacts:
\texttt{traits\_ai\_v1.csv} (5,489,881 rows, ≈620 MB gzipped) containing
the trait fact table; \texttt{traits\_ai\_runs\_v1.csv} (706,220 rows,
≈180 MB gzipped) containing the per-run telemetry;
\texttt{trait\_registry\_v2.json} capturing the 39-key registry
snapshot; \texttt{audit\_sample\_v0.3.csv} (400-row stratified audit
sample); \texttt{audit\_sample\_v0.3\_rated\_first50.csv} (the 50
author-rated red-zone rows from v0.3);
\texttt{audit\_sample\_quote\_supports\_v0.4.csv} (100-row v0.4
quote-supports-value audit, blank and rated versions); and
\texttt{manifest.json} carrying SHA-256 hashes and row counts. Parallel
JSON-LD NDJSON distributions accompany the CSVs; a Darwin Core Archive
descriptor (\texttt{meta.xml}) registers the two extension files and
their field mappings against the TDWG Darwin Core {[}13{]} schema,
attaching the trait layer to \texttt{core\_taxon} by \texttt{taxonID} (=
\texttt{species\_id}). Text data is released under CC-BY 4.0.
Programmatic access is available through the REST API at
\texttt{https://tropicals.cn/api/v1/*}; the species detail endpoint
(\texttt{/api/v1/species/\{slug\}}) returns the trait set inline
together with evidence quotes and registry metadata, and supports
paginated retrieval and per-trait filtering.

\hypertarget{technical-validation}{%
\section{4 Technical Validation}\label{technical-validation}}

This section reports the validation evidence available for the deposit,
ordered by strength. Three orthogonal layers each bound a different
question (see Figure 5 for the overall layer schema). \textbf{Layer 1
(quote-provenance):} substring verification at full population (90.12\%
/ 93.49\%) --- §4.2. \textbf{Layer 2 (quote-supports-value):} semantic
grounding at n=100 single-author audit (100/100 supports, two-sided 95\%
Wilson {[}14{]} lower bound 96.30\%) --- §4.5.2. \textbf{Layer 3
(face-validity):} red-zone subset at n=50 single-author audit (50/50
Accept, two-sided 95\% Wilson lower bound 92.86\%) --- §4.5.1.
Per-record correctness at deposit scale (\texttt{admin\_review\_status}
100\% pending) is a separate non-validation note, not a fourth layer.
Together the three layers bound the framework-level error rate across
three independent questions. §4.6 catalogues what this paper does not
validate.

\begin{figure}
\centering
\includegraphics{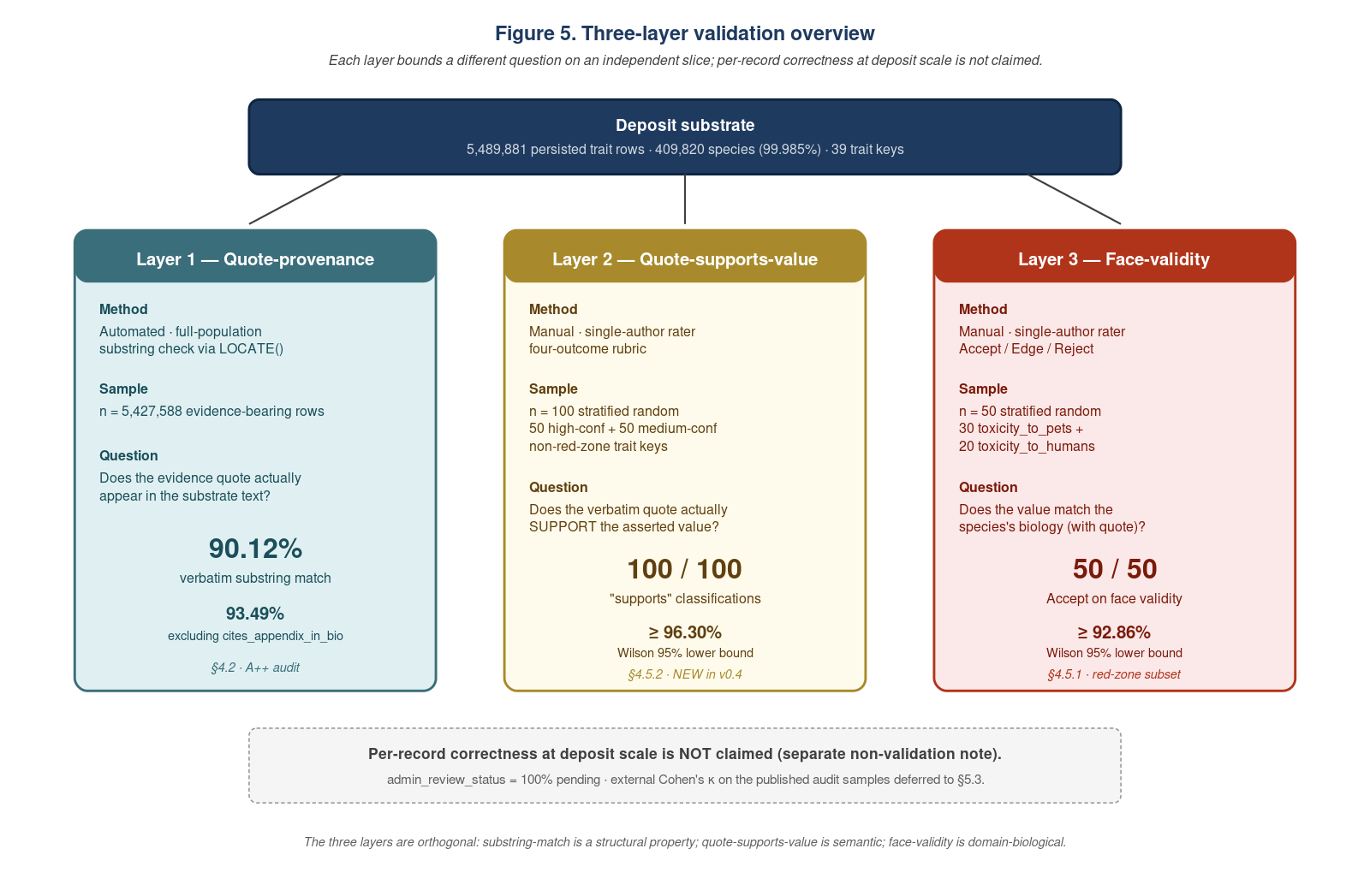}
\caption{Three-layer validation overview. Layer 1 (substring) is
automated at full population; Layers 2-3 are manual single-author
preliminary audits at n=100 and n=50. See §4.5.}
\end{figure}

\textbf{Figure 5.} Three-layer validation schema for the deposit.

\hypertarget{schema-and-registry-conformance-clean-filter-decomposition}{%
\subsection{4.1 Schema and registry conformance --- clean filter
decomposition}\label{schema-and-registry-conformance-clean-filter-decomposition}}

By construction of the pipeline gates documented in §2.4, every
persisted row passes the registry-OOV check, the value-type check, and
the red-zone routing decision. Schema conformance is therefore not an
empirical question for the deposit --- it is a design invariant whose
enforcement we report through the pre-persistence counters of the
706,220 extraction runs. The model emitted approximately 10.84 million
candidate trait cells in total; see Table 2 in §2.4 for the four-way
decomposition into admitted (5,745,167; 52.99\%), substring-filter
rejects (1,025,226; 9.46\%), enum / registry-OOV rejects (9,546;
0.09\%), and model abstentions (4,062,439; 37.47\%).

Three points merit careful reading. First, the substring rejections and
the enum rejections are \textbf{different filter classes} and are
reported separately here; collapsing them into a single ``registry-OOV''
count would hide the dominant grounding mechanism. The substring filter
(9.46\%) rejects candidates whose \texttt{evidence\_quote} does not
appear verbatim in the substrate text; the enum filter (0.09\%) rejects
candidates whose \texttt{trait\_key} is not in the 39-key registry or
whose \texttt{value} is not in the allowed set for a
constrained-vocabulary trait. These are independent mechanisms that
should be counted independently.

Second, \textbf{model abstention (37.47\%) is the dominant non-admission
path} --- far larger than either rejection class. The extractor leaving
a value blank when it cannot extract is a feature of the
registry-constrained protocol (§2.4), not a quality failure: the prompt
explicitly instructs the model to return \texttt{null} rather than
guess. A high abstention rate is what an appropriately calibrated
extractor should produce when the substrate genuinely lacks evidence for
many of the 39 keys per species.

Third, of the 5,745,167 admitted cells, 5,489,881 were persisted; the
255,286-row difference is absorbed by the
\texttt{(species\_id,\ trait\_key,\ model\_version)} UNIQUE constraint.
Across the 433,716 successful runs over 409,820 species (≈24K species
received more than one successful run --- typically retry sessions or
pipeline-validation re-extractions), repeat runs re-emit the same
\texttt{(species\_id,\ trait\_key,\ model\_version)} cells; the UNIQUE
constraint absorbs the second and subsequent emissions, accounting for
the 255,286 admitted-but-not-persisted rows.

Confidence is distributed 81.57\% high and 18.43\% medium across the
5,489,881 persisted rows; low-confidence candidates are dropped at
extraction (§2.4). The red-zone subset --- the four safety-critical
trait keys \texttt{toxicity\_to\_humans}, \texttt{toxicity\_to\_pets},
\texttt{physical\_hazards}, and \texttt{cites\_appendix\_in\_bio} ---
covers 428,668 rows, of which \textbf{376,468 are high-confidence and
52,200 medium, for a high-confidence share of 87.82\%}. This is 6.25
percentage points above the overall 81.57\% rate. The gap is structural
rather than marginal: red-zone keys are dominated by short, well-defined
allowed-value vocabularies (\texttt{physical\_hazards} is a closed
\texttt{multi\_enum}; \texttt{cites\_appendix\_in\_bio} is a four-value
enum), on which the model can be more confident than on traits requiring
inference. The reported figures are confirmed by the Q4 cross-tabulation
in §S5.

\hypertarget{quote-provenance-substring-verification-at-scale}{%
\subsection{4.2 Quote-provenance --- substring verification at
scale}\label{quote-provenance-substring-verification-at-scale}}

Substring verification grounds the \textbf{quote} in the substrate text.
It does not prove that the quote supports the asserted value, nor that
the value itself is correct. The quote-supports-value question is
addressed separately in §4.5.2; per-record correctness is out of scope
(§4.6). With that boundary stated, this is the strongest automated
evidence in the paper: full-population, per-row, addressing the hop-2
hallucination question directly.

Of the 5,489,881 persisted trait rows, 5,427,588 (98.87\%) carry a
non-empty \texttt{evidence\_quote}. For each such row, we test whether
the quote is a verbatim substring of the source
\texttt{species\_detail.bio\_sections} JSON for that species, evaluated
as
\texttt{LOCATE(t.evidence\_quote,\ CAST(sd.bio\_sections\ AS\ CHAR))\ \textgreater{}\ 0}
against the production database. The headline query and its per-key
counterpart each ran for approximately eleven minutes as background
jobs; the documented \texttt{id\ MOD\ 50\ =\ 0} sampling fallback was
not required.

At population level, \textbf{4,891,357 of the 5,427,588 evidence-bearing
rows (90.12\%) verified as verbatim substrings}. This is a conservative
lower bound on hop-2 grounding, for two structural reasons intrinsic to
the test rather than to the extractor. First, roughly 246,041 rows carry
\texttt{evidence\_section\ =\ \textquotesingle{}quick\_card\textquotesingle{}},
indicating the extractor read its evidence from compliance and
quick-card fields (e.g.~the structured \texttt{cites\_status} field)
rather than from the seven-section bio prose. These quotes cannot be
substrings of \texttt{bio\_sections} even when genuinely present and
correctly grounded --- they were never written to that column. Second,
\texttt{CAST(bio\_sections\ AS\ CHAR)} escapes embedded \texttt{"} as
\texttt{\textbackslash{}"}, so a quote containing \texttt{"} will fail a
raw \texttt{LOCATE} match even when the underlying text contains the
quoted phrase.

The single compliance meta-trait \texttt{cites\_appendix\_in\_bio}
accounts for 198,840 of the 536,231 unverified rows (37.1\% of all
misses) --- exactly because its evidence by design references the
structured \texttt{cites\_status} field rather than the bio prose.
\textbf{Excluding \texttt{cites\_appendix\_in\_bio}, substring
verification rises to 93.49\% (4,841,014 of 5,178,405).} Both figures
(90.12\% raw, 93.49\% excluding the compliance meta-trait) are reported
as the headline result.

Per trait key, the picture is uniform and strong. Of the 39 trait keys,
37 verify at ≥80\% and 29 at ≥90\%, with a median per-key rate of
approximately 94\% (Figure 6 shows the full distribution across all 39
keys). Table 3 reports the four highest- and four lowest-verifying keys;
the full per-key table is in §S5.

\begin{figure}
\centering
\includegraphics{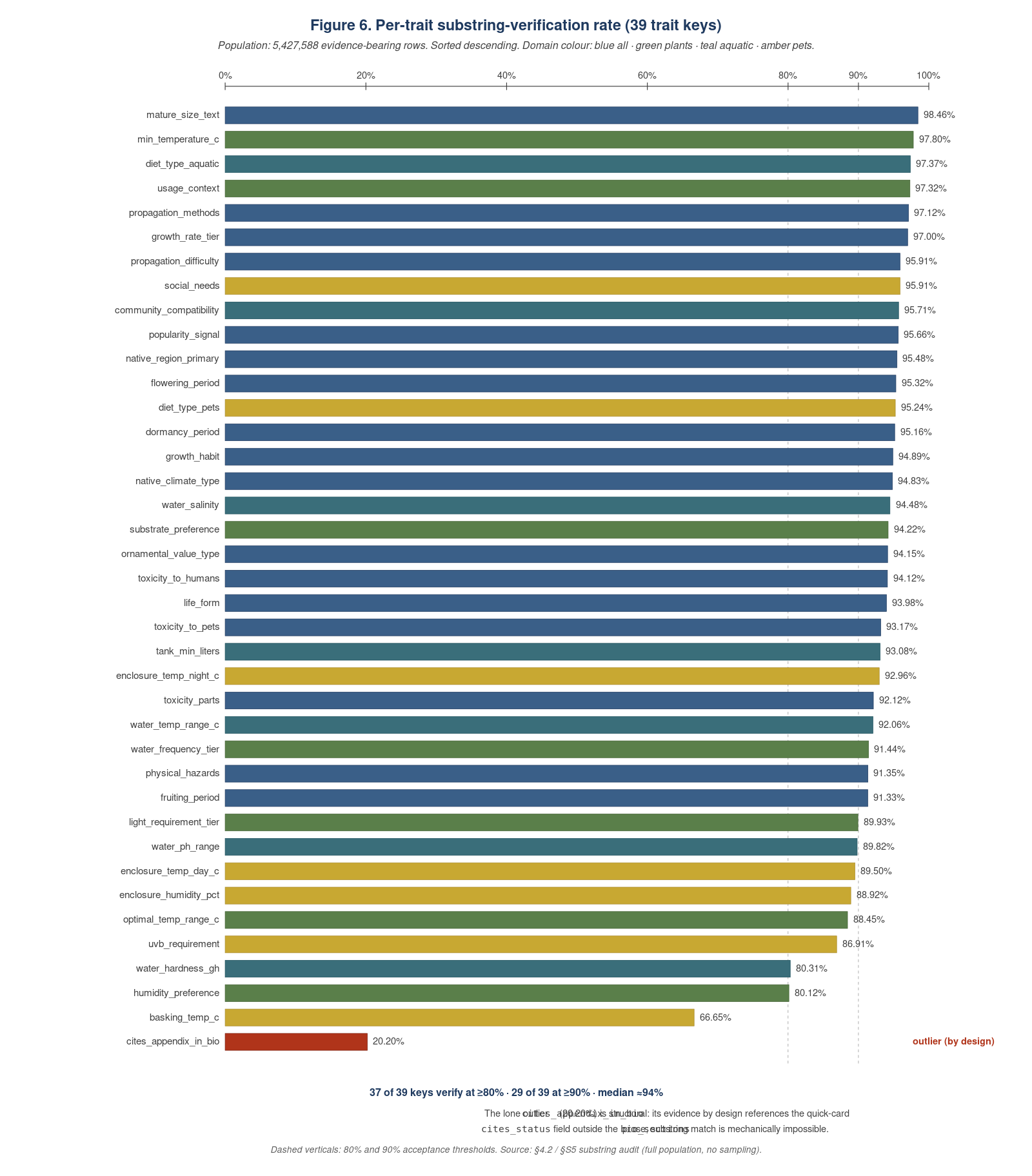}
\caption{Per-trait substring-verification rate (39 keys), sorted
descending. Population: 5,427,588 evidence-bearing rows. Outlier
\texttt{cites\_appendix\_in\_bio} (20.20\%) references quick-card fields
outside \texttt{bio\_sections} by design. Median ≈94\%.}
\end{figure}

\textbf{Figure 6.} Per-trait\_key substring-verification rate across all
39 trait keys.

\textbf{Table 3.} Per-trait\_key substring-verification rate (selected).
Denominators are evidence-quote-bearing rows joined to
\texttt{species\_detail}.

\begin{longtable}[]{@{}lrrr@{}}
\toprule
trait\_key & evidence-bearing rows & substring-verified & pct
verified\tabularnewline
\midrule
\endhead
\texttt{mature\_size\_text} & 244,609 & 240,852 & 98.46\tabularnewline
\texttt{min\_temperature\_c} & 190,563 & 186,380 & 97.80\tabularnewline
\texttt{usage\_context} & 176,294 & 171,566 & 97.32\tabularnewline
\texttt{propagation\_methods} & 251,541 & 244,293 & 97.12\tabularnewline
\ldots{} (37 of 39 keys ≥80\%; see §S5 for full table) & &
&\tabularnewline
\texttt{water\_hardness\_gh} & 20,186 & 16,212 & 80.31\tabularnewline
\texttt{humidity\_preference} & 222,277 & 178,079 & 80.12\tabularnewline
\texttt{basking\_temp\_c} & 12,110 & 8,071 & 66.65\tabularnewline
\texttt{cites\_appendix\_in\_bio} & 249,183 & 50,343 &
20.20\tabularnewline
\bottomrule
\end{longtable}

The single outlier --- \texttt{cites\_appendix\_in\_bio} at 20.20\% ---
is by design. The remaining low-verifying keys
(\texttt{basking\_temp\_c}, \texttt{humidity\_preference},
\texttt{water\_hardness\_gh}) are dominated by quick-card-sourced
evidence and JSON-quote-escape effects rather than by extractor failure;
the high-verifying keys are objective free-text or numeric descriptors
the model can quote directly from the bio prose.

\hypertarget{confidence-distribution-by-trait-type}{%
\subsection{4.3 Confidence distribution by trait
type}\label{confidence-distribution-by-trait-type}}

This section is descriptive, not a calibration claim. Framing this as
evidence that high-confidence rows are more accurate than
medium-confidence ones would overstate what the data establish: we
report the \textbf{distribution} of confidence labels by trait type, not
whether high-confidence rows are in fact more accurate than
medium-confidence rows. A proper calibration study (comparing each
confidence stratum to ground truth on a stratified sample) is deferred
(§5.3). Stratification by confidence is possible for downstream
consumers, but whether it improves correctness on this deposit is not
yet measured.

The descriptive pattern is consistent with what an appropriately
calibrated extractor should produce: measurable numeric and range traits
cluster at the top of the high-confidence share, and inferential or
subjective traits cluster at the bottom. Table 4 reports the five
highest- and five lowest-confidence trait keys; the full distribution is
in §S5.

\textbf{Table 4.} Per-trait\_key model-assigned high-confidence share
(selected). Persisted rows.

\begin{longtable}[]{@{}lrrrr@{}}
\toprule
trait\_key & persisted & high & medium & pct high\tabularnewline
\midrule
\endhead
\texttt{water\_temp\_range\_c} & 47,391 & 47,337 & 54 &
99.89\tabularnewline
\texttt{optimal\_temp\_range\_c} & 268,682 & 268,242 & 440 &
99.84\tabularnewline
\texttt{enclosure\_humidity\_pct} & 87,734 & 87,564 & 170 &
99.81\tabularnewline
\texttt{water\_ph\_range} & 46,723 & 46,592 & 131 & 99.72\tabularnewline
\texttt{native\_region\_primary} & 396,774 & 392,616 & 4,158 &
98.95\tabularnewline
\ldots{} (see §S5 for full table) & & & &\tabularnewline
\texttt{toxicity\_to\_humans} & 12,435 & 6,232 & 6,203 &
50.12\tabularnewline
\texttt{social\_needs} & 22,898 & 9,839 & 13,059 & 42.97\tabularnewline
\texttt{toxicity\_to\_pets} & 6,690 & 2,774 & 3,916 &
41.46\tabularnewline
\texttt{growth\_rate\_tier} & 138,194 & 56,107 & 82,087 &
40.60\tabularnewline
\texttt{dormancy\_period} & 110,359 & 41,468 & 68,891 &
37.58\tabularnewline
\bottomrule
\end{longtable}

Measurable numeric ranges that can be quoted directly from substrate
text (water temperature, pH, enclosure humidity) score at the top; trait
keys requiring inference rather than direct readout (dormancy period,
growth rate tier, social needs, pet toxicity) score at the bottom. The
model is reliably less certain on traits where uncertainty is
appropriate --- which is what an informative confidence signal looks
like, even before a calibration study has been performed.

\hypertarget{multi-version-divergence-opportunity-sample}{%
\subsection{4.4 Multi-version divergence --- opportunity
sample}\label{multi-version-divergence-opportunity-sample}}

The \texttt{(species\_id,\ trait\_key,\ model\_version)} UNIQUE
constraint preserves cross-model history rather than collapsing it. Four
\texttt{model\_version} labels are present at this snapshot: the
canonical \texttt{full-v1-20260524} (5,487,270 rows, 99.95\% of the
deposit), the pilot \texttt{pilot-v1-20260524-1304} (2,431 rows), a
cross-family spot-check \texttt{deepseek-v4-flash-fix-20260529} (157
rows across 27 trait keys), and a smoke-test residue
\texttt{smoke-test-130109} (23 rows). 2,064
\texttt{(species\_id,\ trait\_key)} pairs appear in more than one
model\_version; 546 of those diverge in value, for a value-agreement
rate of 73.5\%.

\textbf{The 2,064-pair set is an opportunity sample, not a designed
cross-model experiment.} Of the 2,064 pairs, \textbf{1,915 (92.8\%) are
within the same mimo family} (combinations of \texttt{full-v1},
\texttt{pilot-v1}, and the smoke-test residue, all runs of the same
underlying model). Only \textbf{149 pairs (7.2\%) are genuinely
cross-family}, involving the \texttt{deepseek-v4-flash-fix-20260529}
batch. The 73.5\% agreement figure therefore primarily measures
same-family run-to-run stability, not cross-family validation. Framing
this as a cross-model agreement signal would overstate what the data
support, and it is reported here only as an opportunity sample. A proper
cross-family study --- re-running the canonical extractor against an
independent LLM family on a stratified sample --- is deferred (§5.3).
The 157-row deepseek batch nevertheless provides a small preliminary
cross-family signal: real but not systematic.

The character of the 546 divergences, with one minor exception, is
cosmetic rather than contradictory. Table 5 stratifies them by
\texttt{value\_type} using distinct-pair denominators.

\textbf{Table 5.} Divergence stratification by \texttt{value\_type}
across the 546 distinct multi-version divergent pairs.

\begin{longtable}[]{@{}lrr@{}}
\toprule
value\_type & divergent pairs & share\tabularnewline
\midrule
\endhead
\texttt{multi\_enum} & 277 & 50.7\%\tabularnewline
\texttt{text} & 178 & 32.6\%\tabularnewline
\texttt{enum} & 86 & 15.7\%\tabularnewline
\texttt{range} & 4 & 0.7\%\tabularnewline
\texttt{int} & 1 & 0.2\%\tabularnewline
\bottomrule
\end{longtable}

\textbf{455 of 546 (83.3\%) of divergences are soft} ---
\texttt{multi\_enum} element ordering or subset differences, and
\texttt{text} paraphrase. Two worked examples:
\texttt{ornamental\_value\_type} for \emph{Microsorum pteropus}
`Windelov' differs as \texttt{form} vs \texttt{foliage,form} (a
multi\_enum subset variation); \texttt{popularity\_signal} for
\emph{Echinodorus} `Red Melon' differs as \texttt{common} vs
\texttt{very\_common} --- an adjacent-level enum disagreement and the
only category that materially changes downstream stratification. Only
\textbf{91 of 546 (16.7\%)} are in constrained-vocabulary fields
(\texttt{enum} 86, \texttt{range} 4, \texttt{int} 1) where the
divergence represents an actual categorical disagreement.

\hypertarget{single-author-preliminary-review-face-validity-and-quote-supports-value}{%
\subsection{4.5 Single-author preliminary review --- face validity and
quote-supports-value}\label{single-author-preliminary-review-face-validity-and-quote-supports-value}}

We report two manual audits, each addressing a question the automated
§4.2 substring test cannot answer. §4.5.1 asks whether values match
species biology on the highest-stakes subset. §4.5.2 --- \textbf{the new
v0.4 result} --- asks whether evidence quotes actually support the
asserted values, bridging quote-existence (§4.2) and per-record
correctness (out of scope, §4.6). Both audits are single-rater
preliminary reviews by the paper author, with samples published for
independent re-rating and Cohen's κ computation.

\hypertarget{face-validity-on-the-red-zone-subset-n-50}{%
\subsubsection{4.5.1 Face validity on the red-zone subset (n =
50)}\label{face-validity-on-the-red-zone-subset-n-50}}

A stratified random sample of 50 rows from canonical
\texttt{full-v1-20260524} covered two of the four red-zone trait keys:
30 rows from \texttt{toxicity\_to\_pets} and 20 from
\texttt{toxicity\_to\_humans}. (Note: only two of the four red-zone keys
--- \texttt{physical\_hazards} and \texttt{cites\_appendix\_in\_bio}
excluded --- are covered in this preliminary subset; see §S6.) The
author classified each row on face-validity criteria (does the asserted
\texttt{value} match the species's biology, and does the
\texttt{evidence\_quote} support it) with three outcomes: Accept, Edge,
Reject. \textbf{All 50 rows were classified Accept; two-sided 95\%
Wilson lower bound (equivalent to one-sided 97.5\%) 92.86\%} (point
estimate 1.000, n = 50).

Caveats are explicit: single non-independent rater (the paper author);
two of four red-zone keys; n = 50 yields a wide interval and is best
understood as a sanity check on the highest-stakes records, not a
precision estimate.

\hypertarget{quote-supports-value-audit-n-100-new-in-v0.4}{%
\subsubsection{4.5.2 Quote-supports-value audit (n = 100) --- new in
v0.4}\label{quote-supports-value-audit-n-100-new-in-v0.4}}

This audit addresses the intermediate question between quote-provenance
(§4.2) and per-record correctness (out of scope, §4.6): given that the
quote exists in source text, does it actually support the value the
extractor emitted?

A stratified random sample of 100 rows was drawn from canonical
\texttt{full-v1-20260524}, balanced as \textbf{50 high-confidence + 50
medium-confidence} rows, and explicitly \textbf{excluding the four
red-zone trait keys} (covered separately in §4.5.1). The sample spans 23
high-confidence and 17 medium-confidence stratum-distinct trait keys
drawn from the 35 non-red-zone keys (the two strata partially overlap).
The author answered a different question than §4.5.1: \textbf{does the
verbatim evidence quote actually support the asserted value?} Four
outcomes were available: supports, partial, does\_not\_support,
cannot\_judge.

\textbf{The result is 100/100 supports; two-sided 95\% Wilson lower
bound (equivalent to one-sided 97.5\%) 96.30\%} (point estimate 1.000, n
= 100). This is the first measurement at non-trivial scale bridging
substring presence (§4.2) and semantic support, and it bounds the
framework-level rate at which evidence quotes --- when they exist
verbatim in the source --- actually justify the values they were emitted
to support.

Caveats parallel §4.5.1: single non-independent rater; n = 100 is a
sanity-check sample with a wide lower bound; non-red-zone only; balanced
on confidence stratum but opportunistic in trait-key spread. Both audit
samples --- \texttt{audit\_sample\_v0.3.csv} (face validity) and
\texttt{audit\_sample\_quote\_supports\_v0.4.csv} plus the author-rated
\texttt{\ldots{}\_rated} companion (quote-supports-value) --- are
published with the deposit. Any independent rater can re-rate without
coordination overhead and compute Cohen's κ against the author's
classifications.

\hypertarget{joint-interpretation-of-4.2-4.5.1-and-4.5.2}{%
\subsubsection{4.5.3 Joint interpretation of §4.2, §4.5.1, and
§4.5.2}\label{joint-interpretation-of-4.2-4.5.1-and-4.5.2}}

The three validation layers each bound a different question on an
independent slice and should not be confused with one another.
\textbf{Layer 1, quote exists in source} (§4.2, automated, full
population): 90.12\% raw / 93.49\% excluding the compliance meta-trait,
at 5.5 million rows. \textbf{Layer 2, quote supports value} (§4.5.2,
manual, n = 100): 100/100, two-sided 95\% Wilson lower bound (equivalent
to one-sided 97.5\%) 96.30\%, single rater, non-red-zone. \textbf{Layer
3, value matches biology} (§4.5.1, manual, n = 50): 50/50, two-sided
95\% Wilson lower bound 92.86\%, single rater, red-zone-only. None of
these establishes per-record correctness at deposit scale. Together they
bound the framework-level error rate across three independent questions;
the published samples make the deferred external Cohen's κ tractable for
any reviewer who wishes to compute it.

\hypertarget{what-this-paper-does-not-validate}{%
\subsection{4.6 What this paper does not
validate}\label{what-this-paper-does-not-validate}}

Five claims are out of scope at this release.

\begin{itemize}
\tightlist
\item
  \textbf{Per-record correctness for any individual trait.}
  \texttt{admin\_review\_status} is 100\% \texttt{pending}; the curation
  infrastructure is published but unexercised. The deposit is positioned
  as an audit-ready substrate, not a per-record-validated trait
  inventory.
\item
  \textbf{Hop-1 fidelity.} The substrate
  \texttt{species\_detail.bio\_sections} text was produced by an
  upstream Qwen-family LLM (P1's pipeline; Wang {[}2{]}). The audits in
  §4.2 and §4.5.2 ground hop 2 (extractor reads and uses its evidence
  accurately); they do not speak to hop 1 (substrate text is faithful to
  its sources). Hop-1 validation is the substrate paper's scope.
\item
  \textbf{Cohen's κ inter-rater agreement.} Not computed here. The
  published audit samples (\texttt{audit\_sample\_v0.3.csv},
  \texttt{audit\_sample\_quote\_supports\_v0.4.csv}) make external blind
  re-rating tractable.
\item
  \textbf{Domain-expert review against external authoritative sources.}
  Only the preliminary spot-check via §4.5.1; full review against ASPCA
  {[}9{]}, Species+ {[}10{]}, IUCN Red List {[}11{]}, POWO {[}12{]} and
  equivalents is deferred.
\item
  \textbf{Cross-family agreement at full scale.} Only the 157-row
  \texttt{deepseek-v4-flash-fix-20260529} opportunity sample provides a
  cross-family signal at this snapshot (§4.4). A properly stratified
  cross-family study is deferred.
\end{itemize}

These deferrals are the legitimate scope boundary of a methodology paper
that publishes the framework and the framework-level evidence. The
provenance machinery (registry, evidence quote, confidence label,
model\_version preservation) is what makes the deferred audits efficient
when they are performed.

\hypertarget{limitations-and-future-work}{%
\section{5 Limitations and Future
Work}\label{limitations-and-future-work}}

\hypertarget{what-this-deposit-does-and-does-not-establish}{%
\subsection{5.1 What this deposit does and does not
establish}\label{what-this-deposit-does-and-does-not-establish}}

The deposit establishes that a registry-bound, evidence-grounded
extraction pipeline can be operated at scale (5,489,881 trait rows
across 409,820 species under a single canonical extractor checkpoint)
with the per-row provenance machinery --- \texttt{evidence\_quote},
\texttt{evidence\_section}, \texttt{ai\_reasoning},
\texttt{model\_version}, \texttt{schema\_version},
\texttt{admin\_review\_status}, \texttt{confidence} --- necessary for
downstream audit. It does not establish per-record correctness for any
single trait row.

Three validation layers, in descending evidentiary strength.
\textbf{Layer 1, quote-provenance}, verified at full population scale:
90.12\% of 5,427,588 evidence-bearing rows have their
\texttt{evidence\_quote} as a verbatim substring of the source
\texttt{bio\_sections} paragraph (93.49\% excluding one compliance-meta
trait whose evidence by design references quick-card fields outside the
prose); the \emph{hop-2} model-to-substrate claim rests on a
deterministic SQL check, not on human judgement. \textbf{Layer 2,
quote-supports-value}, verified at n=100: a single-author sanity check
(50 high + 50 medium rows spanning 23 high-confidence and 17
medium-confidence stratum-distinct trait keys drawn from the 35
non-red-zone keys, under
\texttt{model\_version=\textquotesingle{}full-v1-20260524\textquotesingle{}};
the two strata partially overlap) classified 100/100 rows as supporting
their asserted value (two-sided 95\% Wilson {[}14{]} lower bound,
equivalent to one-sided 97.5\%, 96.30\% on this preliminary sample); a
transparency signal, not a deposit-wide precision claim. \textbf{Layer
3, face-validity on the red-zone subset}, verified at n=50: 50/50 Accept
(two-sided 95\% Wilson lower bound 92.86\%) on a single-author audit
covering two of the four red-zone trait keys; details in §4.5.1.
Per-record correctness at deposit scale is not verified and is a
separate non-validation note, not a fourth layer --- all 5,489,881 rows
carry
\texttt{admin\_review\_status\ =\ \textquotesingle{}pending\textquotesingle{}}.
The four-valued review enum, the composite index
\texttt{idx\_review\_queue\ (admin\_review\_status,\ trait\_key)}, the
red-zone mapping, and the \texttt{/admin/traits/} moderation UI are in
place; the human curator sweep has not been run. The deposit should be
treated as a high-coverage, audit-ready substrate for independent
verification, not a per-record-validated trait inventory.

\hypertarget{known-weaknesses}{%
\subsection{5.2 Known weaknesses}\label{known-weaknesses}}

\textbf{Curation infrastructure published but unexercised.} All
5,489,881 rows are \texttt{pending}; zero are \texttt{approved},
\texttt{rejected}, or \texttt{flagged}. Schema, queue index, and
moderation UI exist; human curator outcomes do not.

\textbf{Single-rater face-validity audit.} The n=100
quote-supports-value figure (§4) is rated by the first author in a
single pass without blinding to the pipeline's design. It characterises
the recommended-consumption subset (high-confidence, non-red-zone,
canonical model version), not the deposit at large. The sample is
published in full so a third party can extend or replicate the review.

\textbf{Hop-2 grounding only.} The substring-verification rate measures
model-to-substrate fidelity. Whether the substrate paragraph itself is
faithful to upstream authoritative literature is \emph{hop-1} fidelity
and is the responsibility of the P1 substrate deposit (Wang {[}2{]}).

\textbf{Substring-verification rate is a conservative floor.} The
90.12\% headline has two structural underestimation sources independent
of extraction quality: (i) \texttt{cites\_appendix\_in\_bio} draws its
evidence from the quick-card / compliance field rather than from a
\texttt{bio\_sections} paragraph (excluding it raises the rate to
93.49\%); (ii) \texttt{CAST(bio\_sections\ AS\ CHAR)} JSON-escapes
embedded ASCII \texttt{"}, so quotes containing a double-quote fail a
raw \texttt{LOCATE} even when genuinely present. The true rate is higher
than 93.49\%; 90.12\% is the most conservative figure computable over
the full deposit without per-trait branching.

\textbf{Failed-run rate is operational, not extraction-quality.} 272,435
of 706,220 logged runs (38.58\%) carry
\texttt{status\ =\ \textquotesingle{}failed\textquotesingle{}},
originating from transient HTTP 502 / 429 / timeout responses on the
upstream model endpoint under cluster-scale concurrent load. Failed runs
return no candidate values and persist no rows; the orchestrator retries
until \texttt{success} or \texttt{filtered\_all}. The 99.985\%
per-species coverage (409,820 of 409,880 publishable species) is direct
evidence that the retry loop closes.

\textbf{Domain coverage bias.} Coverage is biased toward horticulturally
and commercially relevant taxa --- Araceae, Orchidaceae, and related
families are prominent; species with little or no horticultural /
aquarium / herpetoculture footprint are under-represented even at the
substrate level. Downstream analyses sensitive to taxonomic balance
should weight or stratify accordingly.

\textbf{Single dominant \texttt{model\_version} and within-family
cross-model evidence.} 5,487,270 of 5,489,881 rows (99.95\%) were
extracted under \texttt{full-v1-20260524}. The 2,064 multi-version
\texttt{(species\_id,\ trait\_key)} pairs available for cross-version
comparison are overwhelmingly within the same model family (1,915 pairs,
92.8\%, of which the single \texttt{full-v1} vs \texttt{pilot-v1-1304}
combination alone contributes 1,894); only 149 pairs (7.2\%) involve a
different model family (\texttt{deepseek-v4-flash-fix-20260529}).
Run-to-run within-family stability is well sampled; cross-family
agreement is observed only on an opportunity sample and a controlled
comparison is deferred (§5.3).

\hypertarget{principal-deferred-validation}{%
\subsection{5.3 Principal deferred
validation}\label{principal-deferred-validation}}

Four external validations are planned for a subsequent release and are
explicitly not claimed by this deposit:

\textbf{Blind external inter-rater review with Cohen's κ} across the
published v0.3 (400 rows, 120 red-zone + 280 non-red-zone) and v0.4 (100
rows, non-red-zone) audit samples, under a unified Accept / Edge /
Reject rubric. The current deposit reports no κ value.

\textbf{Domain-expert review against authoritative external sources} for
the red-zone trait values (\texttt{toxicity\_to\_pets},
\texttt{toxicity\_to\_humans}, \texttt{physical\_hazards},
\texttt{cites\_appendix\_in\_bio}), spot-checked against ASPCA {[}9{]}
Toxic and Non-Toxic Plants, CITES Species+ {[}10{]}, the IUCN Red List
{[}11{]}, and Kew POWO {[}12{]} on a stratified sample.

\textbf{Stratified cross-family comparison at controlled scale.} The 149
cross-family pairs available in this deposit are a preliminary signal,
not a designed experiment. A planned controlled comparison will re-run
the canonical extraction on a stratified species sample under a non-mimo
LLM family and characterise agreement at the population level.

\textbf{Baseline overlap study against existing curated trait databases}
(TRY {[}3{]}, BIEN {[}4{]}, GIFT {[}5{]}, AusTraits {[}6{]}) on the
intersecting species set, to quantify the deposit's complementarity to
those resources. Deferred pending taxon-key alignment.

\hypertarget{usage-notes}{%
\section{6 Usage Notes}\label{usage-notes}}

\hypertarget{recommended-consumption-patterns}{%
\subsection{6.1 Recommended consumption
patterns}\label{recommended-consumption-patterns}}

Downstream consumers should filter trait rows by the conjunction
\texttt{model\_version\ =\ \textquotesingle{}full-v1-20260524\textquotesingle{}\ AND\ confidence\ =\ \textquotesingle{}high\textquotesingle{}\ AND\ admin\_review\_status\ IN\ (\textquotesingle{}pending\textquotesingle{},\textquotesingle{}approved\textquotesingle{})}
and stratify reporting by \texttt{trait\_key}. The
\texttt{model\_version} pin selects the canonical extraction batch
(99.95\% of the deposit); the \texttt{confidence} filter narrows to the
81.57\% of rows the model self-reported as high-confidence; the
\texttt{admin\_review\_status} clause admits both \texttt{pending} and
\texttt{approved} so the query continues to apply once future releases
add moderator-resolved rows.

Per-trait-key stratification is essential: model-assigned
high-confidence rates range from 99.89\% on
\texttt{water\_temp\_range\_c} (a numeric range typically stated
verbatim) down to 37.58\% on \texttt{dormancy\_period} (an inferential
enum requiring synthesis across sections). A uniform threshold across 39
trait keys is not a substitute for trait-key-aware filtering.

\hypertarget{worked-example-sql-query-for-safe-high-confidence-trait-retrieval}{%
\subsection{6.2 Worked example --- SQL query for safe high-confidence
trait
retrieval}\label{worked-example-sql-query-for-safe-high-confidence-trait-retrieval}}

\begin{Shaded}
\begin{Highlighting}[]
\KeywordTok{SELECT}\NormalTok{ trait\_key, }\FunctionTok{value}\NormalTok{, evidence\_quote, evidence\_section, ai\_reasoning, model\_version}
\KeywordTok{FROM}\NormalTok{ species\_traits\_ai}
\KeywordTok{WHERE}\NormalTok{ species\_id }\OperatorTok{=} \DecValTok{12345}
  \KeywordTok{AND}\NormalTok{ model\_version }\OperatorTok{=} \StringTok{\textquotesingle{}full{-}v1{-}20260524\textquotesingle{}}
  \KeywordTok{AND}\NormalTok{ confidence }\OperatorTok{=} \StringTok{\textquotesingle{}high\textquotesingle{}}
  \KeywordTok{AND}\NormalTok{ admin\_review\_status }\KeywordTok{IN}\NormalTok{ (}\StringTok{\textquotesingle{}pending\textquotesingle{}}\NormalTok{, }\StringTok{\textquotesingle{}approved\textquotesingle{}}\NormalTok{)}
\KeywordTok{ORDER} \KeywordTok{BY}\NormalTok{ trait\_key;}
\end{Highlighting}
\end{Shaded}

Red-zone keys (\texttt{toxicity\_to\_humans},
\texttt{toxicity\_to\_pets}, \texttt{physical\_hazards},
\texttt{cites\_appendix\_in\_bio}) returned by this query should not be
consumed without independent verification against an authoritative
source. The public API serves them inline as informational signals only.

\hypertarget{stratification-by-confidence}{%
\subsection{6.3 Stratification by
confidence}\label{stratification-by-confidence}}

The four red-zone trait keys account for 428,668 rows; their aggregate
high-confidence rate is 87.82\% (376,468 of 428,668), 6.25 percentage
points above the corpus-wide 81.57\% rate. The aggregate masks
substantial per-key variation: rates range from 41.5\%
(\texttt{toxicity\_to\_pets}) to 98.8\%
(\texttt{cites\_appendix\_in\_bio}). Consumers using red-zone rows for
household-safety screening, regulatory-compliance review, or
human-medical decision support must independently verify each row
against an authoritative source (ASPCA {[}9{]}, CITES Species+ {[}10{]},
IUCN Red List {[}11{]}, POWO {[}12{]}). The \texttt{pending} review
status is reported as a deposited fact, not a temporary deficiency.

\hypertarget{data-availability}{%
\section{7 Data Availability}\label{data-availability}}

\hypertarget{substrate}{%
\subsection{7.1 Substrate}\label{substrate}}

The trait-extension deposit attaches to the cross-domain tropical
species substrate \emph{Tropicals.cn: Tropical Species Encyclopedia}
(Wang {[}1{]}, substrate data DOI \texttt{10.5281/zenodo.20377811};
companion Data Descriptor Wang {[}2{]}, DOI
\texttt{10.5281/zenodo.20424981}). The substrate supplies the identifier
graph (\texttt{species} table including the surrogate key
\texttt{species\_id}), the upstream encyclopedia content
(\texttt{species\_detail.bio\_sections}), and the publishable-gate
predicate. Cross-deposit analyses should cite both deposits; the trait
deposit references the substrate via id-only foreign key and never
duplicates upstream descriptive text.

\hypertarget{this-deposit}{%
\subsection{7.2 This deposit}\label{this-deposit}}

The trait-extension release is published as an \textbf{independent
Zenodo record} with its own Concept DOI, issued at the time of first
publication. Subsequent releases are published as new versions under
that Concept DOI through Zenodo's \texttt{newversion} workflow, so the
Concept DOI continues to resolve to the latest version while each
release remains individually citable through its version-specific DOI.
The trait deposit is never published as a new version of the substrate's
Concept DOI; the two records are versioned independently. The deposit's
primary artifacts are enumerated in §3.5: the trait fact table CSV, the
per-run telemetry CSV, the registry JSON snapshot, the v0.3 and v0.4
audit samples (blank and rated), and the SHA-256 manifest, with a Darwin
Core Archive descriptor (\texttt{meta.xml}) registering the trait files
as new extensions linked to the substrate's \texttt{core\_taxon} by
\texttt{taxonID} (= \texttt{species\_id}). Text data is released under
CC-BY 4.0.

\hypertarget{code-availability}{%
\section{8 Code Availability}\label{code-availability}}

The trait-extraction pipeline source code is published alongside the
deposit under CC-BY 4.0. Primary source files are
\texttt{scripts/mimo/extract-traits.mjs} (the production extractor
implementing the registry-conformance gate, source-span filter, and run
telemetry writer) and \texttt{scripts/mimo/trait-registry.mjs} (the
canonical 39-key registry definition). Operational cluster-management
scripts referenced in the supplementary material and the audit-sample
materialisation script are included in the same repository. The public
repository URL will be published alongside the Zenodo deposit and
recorded in the version record's metadata.

\hypertarget{author-contributions}{%
\section{9 Author Contributions}\label{author-contributions}}

J.W. conceived the pipeline architecture, implemented the extraction and
admission pipeline, designed the audit protocols, conducted the
single-author preliminary reviews (§4.5.1 and §4.5.2), performed the
data analyses, and wrote the manuscript.

\hypertarget{competing-interests}{%
\section{10 Competing Interests}\label{competing-interests}}

The author declares no competing interests. The author is the founder of
Tropicals.cn, the substrate from which the trait deposit is extracted;
this affiliation is disclosed for transparency and does not constitute a
competing financial or non-financial interest in the present
methodological contribution.

\hypertarget{funding}{%
\section{11 Funding}\label{funding}}

This work received no external funding. Compute and infrastructure costs
were borne by NEXLY LLC.

\hypertarget{acknowledgements}{%
\section{12 Acknowledgements}\label{acknowledgements}}

The author thanks the maintainers of the upstream catalogues (GBIF
{[}7{]}, iNaturalist {[}8{]}, POWO {[}12{]}, CITES Species+ {[}10{]},
IUCN Red List {[}11{]}) and the curators of the comparison trait
databases (TRY {[}3{]}, BIEN {[}4{]}, GIFT {[}5{]}, AusTraits {[}6{]}),
whose openly published data made the cross-reference design of this
paper feasible. The mimo-v2.5 extractor is provided by Xiaomi AI; usage
falls within the public-API tier terms.

\hypertarget{references}{%
\section{References}\label{references}}

The reference list below is organised by \textbf{topical grouping}
rather than strict first-appearance order: own-substrate work
{[}1--2{]}, comparable trait databases {[}3--6{]}, biodiversity
infrastructure precedents {[}7--8{]}, authoritative regulatory and
conservation sources {[}9--12{]}, data standards {[}13{]}, and
statistical methodology {[}14{]}. This grouping aids reader navigation;
first-appearance order within each topical block is preserved.

{[}1{]} Wang, J. (2026). Tropicals.cn: Tropical Species Encyclopedia
(v1.0.1) {[}Data set{]}. Zenodo. https://doi.org/10.5281/zenodo.20377811

{[}2{]} Wang, J. (2026). A cross-domain tropical species dataset with
Chinese vernacular names and CITES source links {[}Data Descriptor{]}.
Zenodo. https://doi.org/10.5281/zenodo.20424981

{[}3{]} Kattge, J., Bönisch, G., Díaz, S., Lavorel, S., Prentice, I. C.,
Leadley, P., et al.~(2020). TRY plant trait database --- enhanced
coverage and open access. Global Change Biology, 26(1), 119--188.
https://doi.org/10.1111/gcb.14904. Database portal:
https://www.try-db.org

{[}4{]} Maitner, B. S., Boyle, B., Casler, N., Condit, R., Donoghue, J.,
Durán, S. M., et al.~(2018). The bien r package: A tool to access the
Botanical Information and Ecology Network (BIEN) database. Methods in
Ecology and Evolution, 9(2), 373--379.
https://doi.org/10.1111/2041-210X.12861

{[}5{]} Weigelt, P., König, C., \& Kreft, H. (2020). GIFT --- A Global
Inventory of Floras and Traits for macroecology and biogeography.
Journal of Biogeography, 47(1), 16--43.
https://doi.org/10.1111/jbi.13623

{[}6{]} Falster, D., Gallagher, R., Wenk, E. H., Wright, I. J.,
Indiarto, D., Andrew, S. C., et al.~(2021). AusTraits, a curated plant
trait database for the Australian flora. Scientific Data, 8, 254.
https://doi.org/10.1038/s41597-021-01006-6

{[}7{]} GBIF Secretariat (2024). GBIF: The Global Biodiversity
Information Facility. https://www.gbif.org

{[}8{]} iNaturalist (2024). iNaturalist --- A joint initiative of the
California Academy of Sciences and the National Geographic Society.
https://www.inaturalist.org

{[}9{]} American Society for the Prevention of Cruelty to Animals
(ASPCA) (2024). Toxic and Non-Toxic Plants.
https://www.aspca.org/pet-care/animal-poison-control/toxic-and-non-toxic-plants

{[}10{]} UNEP-WCMC and CITES Secretariat (2024). Species+.
https://www.speciesplus.net

{[}11{]} IUCN (2024). The IUCN Red List of Threatened Species.
https://www.iucnredlist.org

{[}12{]} Royal Botanic Gardens, Kew (2024). Plants of the World Online
(POWO). https://powo.science.kew.org

{[}13{]} Wieczorek, J., Bloom, D., Guralnick, R., Blum, S., Döring, M.,
Giovanni, R., Robertson, T., \& Vieglais, D. (2012). Darwin Core: An
evolving community-developed biodiversity data standard. PLoS ONE, 7(1),
e29715. https://doi.org/10.1371/journal.pone.0029715

{[}14{]} Wilson, E. B. (1927). Probable Inference, the Law of
Succession, and Statistical Inference. Journal of the American
Statistical Association, 22(158), 209--212.
https://doi.org/10.1080/01621459.1927.10502953

\hypertarget{supplementary-materials}{%
\section{Supplementary Materials}\label{supplementary-materials}}

This appendix supplies the auxiliary material referenced from the main
text: (S1) the full enumeration of the 39-key trait registry; (S2) the
production CREATE TABLE statements for the two tables comprising the
deposit; (S3) a compact account of pipeline operations and the principal
failure modes; (S4) the full coverage matrix and the
trait-rows-per-species histogram; (S5) the substring-verification
methodology underlying the hop-2 grounding figure and the
quote-supports-value rubric used for the v0.4 face-validity audit; and
(S6) the audit-sample design and rater protocol covering both the v0.3
(400-row) and v0.4 (100-row) published samples. All numbers are
reproduced from the production database snapshot of 2026-05-29 and the
structural registry snapshot \texttt{trait\_registry\_v2.json} produced
from the same workspace state. Source files are cited at first reference
and are bundled with the deposit.

\begin{center}\rule{0.5\linewidth}{0.5pt}\end{center}

\hypertarget{s1-trait-registry-full-39-key-enumeration}{%
\subsection{S1 Trait registry --- full 39-key
enumeration}\label{s1-trait-registry-full-39-key-enumeration}}

The production registry comprises 39 trait keys: 18 universal, 7
plant-specialised, 7 aquatic-specialised, 7 pet-specialised. Five
\texttt{value\_type} classes: 15 \texttt{enum}, 10 \texttt{multi\_enum},
7 \texttt{range}, 4 \texttt{text}, 3 \texttt{int}. Four keys carry the
\texttt{red\_zone} flag (marked \texttt{RZ} below); a fifth canonical
safety-critical key \texttt{iucn\_status} is not in the 39-key
extraction registry and is flagged for v1.x expansion.

Each table below summarises the keys in one domain; allowed-value
vocabularies follow as indented lists. The authoritative
machine-readable specification is \texttt{trait\_registry\_v2.json},
shipped with the deposit.

\hypertarget{universal-traits-applies_to-all-18-keys}{%
\subsubsection{\texorpdfstring{Universal traits
(\texttt{applies\_to\ =\ all}, 18
keys)}{Universal traits (applies\_to = all, 18 keys)}}\label{universal-traits-applies_to-all-18-keys}}

\begin{longtable}[]{@{}llrc@{}}
\toprule
trait\_key & type & n\_values & red\tabularnewline
\midrule
\endhead
\texttt{mature\_size\_text} & text & --- & ---\tabularnewline
\texttt{growth\_rate\_tier} & enum & 5 & ---\tabularnewline
\texttt{growth\_habit} & multi\_enum & 11 & ---\tabularnewline
\texttt{life\_form} & enum & 5 & ---\tabularnewline
\texttt{native\_region\_primary} & text & --- & ---\tabularnewline
\texttt{native\_climate\_type} & multi\_enum & 9 & ---\tabularnewline
\texttt{propagation\_methods} & multi\_enum & 11 & ---\tabularnewline
\texttt{propagation\_difficulty} & enum & 4 & ---\tabularnewline
\texttt{flowering\_period} & text & --- & ---\tabularnewline
\texttt{fruiting\_period} & text & --- & ---\tabularnewline
\texttt{dormancy\_period} & enum & 4 & ---\tabularnewline
\texttt{toxicity\_to\_humans} & enum & 5 & RZ\tabularnewline
\texttt{toxicity\_to\_pets} & enum & 5 & RZ\tabularnewline
\texttt{toxicity\_parts} & multi\_enum & 16 & ---\tabularnewline
\texttt{physical\_hazards} & multi\_enum & 8 & RZ\tabularnewline
\texttt{cites\_appendix\_in\_bio} & enum & 4 & RZ\tabularnewline
\texttt{ornamental\_value\_type} & multi\_enum & 10 & ---\tabularnewline
\texttt{popularity\_signal} & enum & 4 & ---\tabularnewline
\bottomrule
\end{longtable}

\textbf{Allowed-value vocabularies (universal traits):}

\begin{itemize}
\tightlist
\item
  \texttt{mature\_size\_text}: free-form text, ≤80 chars (e.g.~``0.5-1.5
  m height, 30-60 cm spread'')
\item
  \texttt{growth\_rate\_tier}: very\_slow, slow, moderate, fast,
  very\_fast
\item
  \texttt{growth\_habit}: upright, climbing, trailing, rosette,
  clumping, spreading, columnar, prostrate, epiphytic, aquatic,
  terrestrial
\item
  \texttt{life\_form}: annual, biennial, perennial,
  evergreen\_perennial, deciduous\_perennial
\item
  \texttt{native\_region\_primary}: free-form text, ≤60 chars
  (e.g.~``Amazon basin'')
\item
  \texttt{native\_climate\_type}: tropical\_rainforest, tropical\_dry,
  tropical\_savanna, subtropical, montane, desert, swamp, temperate,
  coastal
\item
  \texttt{propagation\_methods}: seed, cutting, division, layering,
  grafting, tissue\_culture, spore, offset, plantlet, egg, live\_birth
\item
  \texttt{propagation\_difficulty}: easy, moderate, difficult,
  very\_difficult
\item
  \texttt{flowering\_period}: free-form text, ≤60 chars
  (e.g.~``May--July'' / ``year-round'')
\item
  \texttt{fruiting\_period}: free-form text, ≤60 chars (e.g.~``autumn'';
  null if no fruit)
\item
  \texttt{dormancy\_period}: winter, summer, brief, none
\item
  \texttt{toxicity\_to\_humans} / \texttt{toxicity\_to\_pets}:
  non\_toxic, mildly\_toxic, moderately\_toxic, highly\_toxic, deadly
\item
  \texttt{toxicity\_parts}: leaves, stems, sap, flowers, fruits, seeds,
  roots, bulbs, all\_parts, fangs, spines, stinger, skin\_secretion,
  venom\_glands, barbs, other
\item
  \texttt{physical\_hazards}: thorns, spines, irritating\_sap,
  allergenic\_pollen, sharp\_edges, bite, venom, none
\item
  \texttt{cites\_appendix\_in\_bio}: I, II, III, not\_mentioned
\item
  \texttt{ornamental\_value\_type}: foliage, flower, fruit, form,
  fragrance, bark, seasonal\_color, whole\_plant, behavior, coloration
\item
  \texttt{popularity\_signal}: rare\_collector, uncommon, common,
  very\_common
\end{itemize}

\hypertarget{plant-specialised-traits-applies_to-plants-7-keys}{%
\subsubsection{\texorpdfstring{Plant-specialised traits
(\texttt{applies\_to\ =\ plants}, 7
keys)}{Plant-specialised traits (applies\_to = plants, 7 keys)}}\label{plant-specialised-traits-applies_to-plants-7-keys}}

\begin{longtable}[]{@{}llrl@{}}
\toprule
trait\_key & type & n\_values & spec\tabularnewline
\midrule
\endhead
\texttt{min\_temperature\_c} & int & --- & range {[}-10, 30{]}
°C\tabularnewline
\texttt{optimal\_temp\_range\_c} & range & --- & ``x-y'' within {[}0,
45{]} °C\tabularnewline
\texttt{light\_requirement\_tier} & enum & 6 & tier (see
vocab)\tabularnewline
\texttt{water\_frequency\_tier} & enum & 5 & tier (see
vocab)\tabularnewline
\texttt{humidity\_preference} & enum & 4 & tier (see
vocab)\tabularnewline
\texttt{substrate\_preference} & multi\_enum & 12 & substrate
set\tabularnewline
\texttt{usage\_context} & multi\_enum & 14 & usage set\tabularnewline
\bottomrule
\end{longtable}

\textbf{Allowed-value vocabularies (plants):}

\begin{itemize}
\tightlist
\item
  \texttt{light\_requirement\_tier}: deep\_shade, shade, partial\_shade,
  bright\_indirect, full\_sun, wide\_tolerance
\item
  \texttt{water\_frequency\_tier}: drought\_tolerant, occasional,
  regular, constant\_moisture, semi\_aquatic
\item
  \texttt{humidity\_preference}: low\_lt40, medium\_40\_60,
  high\_60\_80, very\_high\_gt80
\item
  \texttt{substrate\_preference}: well\_drained, moist, sandy, loamy,
  peaty, mounted, epiphytic, aquatic, acidic, calcareous, organic\_rich,
  coarse
\item
  \texttt{usage\_context}: indoor, outdoor, greenhouse, balcony, garden,
  ground\_cover, hedge, specimen, bonsai, cut\_flower, terrarium,
  paludarium, office, public\_space
\end{itemize}

\hypertarget{aquatic-specialised-traits-applies_to-aquatic-7-keys}{%
\subsubsection{\texorpdfstring{Aquatic-specialised traits
(\texttt{applies\_to\ =\ aquatic}, 7
keys)}{Aquatic-specialised traits (applies\_to = aquatic, 7 keys)}}\label{aquatic-specialised-traits-applies_to-aquatic-7-keys}}

\begin{longtable}[]{@{}llrl@{}}
\toprule
trait\_key & type & n\_values & spec\tabularnewline
\midrule
\endhead
\texttt{water\_temp\_range\_c} & range & --- & ``x-y'' within {[}5,
35{]} °C\tabularnewline
\texttt{water\_ph\_range} & range & --- & ``x-y'' within {[}0, 14{]}
pH\tabularnewline
\texttt{water\_hardness\_gh} & range & --- & ``x-y'' within {[}0, 30{]}
GH\tabularnewline
\texttt{water\_salinity} & enum & 4 & see vocab\tabularnewline
\texttt{tank\_min\_liters} & int & --- & range {[}5, 10000{]}
L\tabularnewline
\texttt{community\_compatibility} & enum & 5 & see vocab\tabularnewline
\texttt{diet\_type\_aquatic} & multi\_enum & 6 & see
vocab\tabularnewline
\bottomrule
\end{longtable}

\textbf{Allowed-value vocabularies (aquatic):}

\begin{itemize}
\tightlist
\item
  \texttt{water\_salinity}: freshwater, brackish, marine, euryhaline
\item
  \texttt{community\_compatibility}: solitary, pairs, small\_group,
  large\_school, community\_tank
\item
  \texttt{diet\_type\_aquatic}: carnivore, herbivore, omnivore,
  planktivore, detritivore, piscivore
\end{itemize}

\hypertarget{pet-specialised-traits-applies_to-pets-7-keys}{%
\subsubsection{\texorpdfstring{Pet-specialised traits
(\texttt{applies\_to\ =\ pets}, 7
keys)}{Pet-specialised traits (applies\_to = pets, 7 keys)}}\label{pet-specialised-traits-applies_to-pets-7-keys}}

\begin{longtable}[]{@{}llrl@{}}
\toprule
trait\_key & type & n\_values & spec\tabularnewline
\midrule
\endhead
\texttt{enclosure\_temp\_day\_c} & range & --- & ``x-y'' within {[}10,
45{]} °C\tabularnewline
\texttt{enclosure\_temp\_night\_c} & range & --- & ``x-y'' within {[}10,
40{]} °C\tabularnewline
\texttt{basking\_temp\_c} & int & --- & range {[}20, 50{]}
°C\tabularnewline
\texttt{uvb\_requirement} & enum & 4 & see vocab\tabularnewline
\texttt{enclosure\_humidity\_pct} & range & --- & ``x-y'' within {[}0,
100{]} \%\tabularnewline
\texttt{diet\_type\_pets} & multi\_enum & 7 & see vocab\tabularnewline
\texttt{social\_needs} & enum & 4 & see vocab\tabularnewline
\bottomrule
\end{longtable}

\textbf{Allowed-value vocabularies (pets):}

\begin{itemize}
\tightlist
\item
  \texttt{uvb\_requirement}: required\_high, required\_low, recommended,
  not\_required
\item
  \texttt{diet\_type\_pets}: insectivore, carnivore, herbivore,
  omnivore, frugivore, piscivore, nectarivore
\item
  \texttt{social\_needs}: solitary, pairs, small\_group, communal
\end{itemize}

Registry source of truth: \texttt{scripts/mimo/trait-registry.mjs}.
Structural snapshot used for this paper:
\texttt{trait\_registry\_v2.json}, dumped 2026-05-29.

\hypertarget{s2-production-sql-ddl-species_traits_ai-species_traits_ai_runs}{%
\subsection{\texorpdfstring{S2 Production SQL DDL
(\texttt{species\_traits\_ai} +
\texttt{species\_traits\_ai\_runs})}{S2 Production SQL DDL (species\_traits\_ai + species\_traits\_ai\_runs)}}\label{s2-production-sql-ddl-species_traits_ai-species_traits_ai_runs}}

Both tables target MySQL 8.0 in local development and MySQL 5.7.44 in
production deployment. The DDL is written to be compatible with both: no
CTEs, no window functions, no \texttt{JSON\_TABLE}, no lateral derived
tables. The statements below are reproduced verbatim from the production
schema snapshot of 2026-05-29; main text §3.1 aligns to this DDL.

\hypertarget{species_traits_ai-the-trait-fact-table}{%
\subsubsection{\texorpdfstring{\texttt{species\_traits\_ai} --- the
trait fact
table}{species\_traits\_ai --- the trait fact table}}\label{species_traits_ai-the-trait-fact-table}}

\begin{Shaded}
\begin{Highlighting}[]
\KeywordTok{CREATE} \KeywordTok{TABLE}\NormalTok{ \textasciigrave{}species\_traits\_ai\textasciigrave{} (}
\NormalTok{  \textasciigrave{}id\textasciigrave{} bigint }\KeywordTok{NOT} \KeywordTok{NULL}\NormalTok{ AUTO\_INCREMENT,}
\NormalTok{  \textasciigrave{}species\_id\textasciigrave{} bigint }\KeywordTok{NOT} \KeywordTok{NULL}\NormalTok{,}
\NormalTok{  \textasciigrave{}trait\_key\textasciigrave{} }\DataTypeTok{varchar}\NormalTok{(}\DecValTok{64}\NormalTok{) COLLATE utf8mb4\_unicode\_ci }\KeywordTok{NOT} \KeywordTok{NULL}\NormalTok{,}
\NormalTok{  \textasciigrave{}value\textasciigrave{} }\DataTypeTok{varchar}\NormalTok{(}\DecValTok{255}\NormalTok{) COLLATE utf8mb4\_unicode\_ci }\KeywordTok{DEFAULT} \KeywordTok{NULL}\NormalTok{,}
\NormalTok{  \textasciigrave{}value\_type\textasciigrave{} enum(}\StringTok{\textquotesingle{}enum\textquotesingle{}}\NormalTok{,}\StringTok{\textquotesingle{}multi\_enum\textquotesingle{}}\NormalTok{,}\StringTok{\textquotesingle{}int\textquotesingle{}}\NormalTok{,}\StringTok{\textquotesingle{}range\textquotesingle{}}\NormalTok{,}\StringTok{\textquotesingle{}text\textquotesingle{}}\NormalTok{,}\StringTok{\textquotesingle{}bool\textquotesingle{}}\NormalTok{) COLLATE utf8mb4\_unicode\_ci }\KeywordTok{NOT} \KeywordTok{NULL}\NormalTok{,}
\NormalTok{  \textasciigrave{}confidence\textasciigrave{} enum(}\StringTok{\textquotesingle{}high\textquotesingle{}}\NormalTok{,}\StringTok{\textquotesingle{}medium\textquotesingle{}}\NormalTok{) COLLATE utf8mb4\_unicode\_ci }\KeywordTok{NOT} \KeywordTok{NULL}\NormalTok{,}
\NormalTok{  \textasciigrave{}evidence\_quote\textasciigrave{} }\DataTypeTok{varchar}\NormalTok{(}\DecValTok{500}\NormalTok{) COLLATE utf8mb4\_unicode\_ci }\KeywordTok{DEFAULT} \KeywordTok{NULL}\NormalTok{,}
\NormalTok{  \textasciigrave{}evidence\_section\textasciigrave{} }\DataTypeTok{varchar}\NormalTok{(}\DecValTok{32}\NormalTok{) COLLATE utf8mb4\_unicode\_ci }\KeywordTok{DEFAULT} \KeywordTok{NULL}\NormalTok{,}
\NormalTok{  \textasciigrave{}ai\_reasoning\textasciigrave{} }\DataTypeTok{varchar}\NormalTok{(}\DecValTok{300}\NormalTok{) COLLATE utf8mb4\_unicode\_ci }\KeywordTok{DEFAULT} \KeywordTok{NULL}\NormalTok{,}
\NormalTok{  \textasciigrave{}model\_version\textasciigrave{} }\DataTypeTok{varchar}\NormalTok{(}\DecValTok{64}\NormalTok{) COLLATE utf8mb4\_unicode\_ci }\KeywordTok{NOT} \KeywordTok{NULL}\NormalTok{,}
\NormalTok{  \textasciigrave{}schema\_version\textasciigrave{} }\DataTypeTok{varchar}\NormalTok{(}\DecValTok{16}\NormalTok{) COLLATE utf8mb4\_unicode\_ci }\KeywordTok{NOT} \KeywordTok{NULL}\NormalTok{,}
\NormalTok{  \textasciigrave{}extracted\_at\textasciigrave{} datetime }\KeywordTok{DEFAULT} \FunctionTok{CURRENT\_TIMESTAMP}\NormalTok{,}
\NormalTok{  \textasciigrave{}admin\_review\_status\textasciigrave{} enum(}\StringTok{\textquotesingle{}pending\textquotesingle{}}\NormalTok{,}\StringTok{\textquotesingle{}approved\textquotesingle{}}\NormalTok{,}\StringTok{\textquotesingle{}rejected\textquotesingle{}}\NormalTok{,}\StringTok{\textquotesingle{}flagged\textquotesingle{}}\NormalTok{) COLLATE utf8mb4\_unicode\_ci }\KeywordTok{DEFAULT} \StringTok{\textquotesingle{}pending\textquotesingle{}}\NormalTok{,}
  \KeywordTok{PRIMARY} \KeywordTok{KEY}\NormalTok{ (\textasciigrave{}id\textasciigrave{}),}
  \KeywordTok{UNIQUE} \KeywordTok{KEY}\NormalTok{ \textasciigrave{}uq\_species\_trait\_version\textasciigrave{} (\textasciigrave{}species\_id\textasciigrave{},\textasciigrave{}trait\_key\textasciigrave{},\textasciigrave{}model\_version\textasciigrave{}),}
  \KeywordTok{KEY}\NormalTok{ \textasciigrave{}idx\_trait\_value\textasciigrave{} (\textasciigrave{}trait\_key\textasciigrave{},\textasciigrave{}value\textasciigrave{}),}
  \KeywordTok{KEY}\NormalTok{ \textasciigrave{}idx\_species\_status\textasciigrave{} (\textasciigrave{}species\_id\textasciigrave{},\textasciigrave{}admin\_review\_status\textasciigrave{}),}
  \KeywordTok{KEY}\NormalTok{ \textasciigrave{}idx\_review\_queue\textasciigrave{} (\textasciigrave{}admin\_review\_status\textasciigrave{},\textasciigrave{}trait\_key\textasciigrave{}),}
  \KeywordTok{KEY}\NormalTok{ \textasciigrave{}idx\_model\_extracted\textasciigrave{} (\textasciigrave{}model\_version\textasciigrave{},\textasciigrave{}extracted\_at\textasciigrave{}),}
  \KeywordTok{KEY}\NormalTok{ \textasciigrave{}idx\_trait\_value\_species\textasciigrave{} (\textasciigrave{}trait\_key\textasciigrave{},\textasciigrave{}value\textasciigrave{},\textasciigrave{}species\_id\textasciigrave{}),}
  \KeywordTok{KEY}\NormalTok{ \textasciigrave{}idx\_review\_species\_trait\textasciigrave{} (\textasciigrave{}admin\_review\_status\textasciigrave{},\textasciigrave{}species\_id\textasciigrave{},\textasciigrave{}trait\_key\textasciigrave{}),}
  \KeywordTok{CONSTRAINT}\NormalTok{ \textasciigrave{}fk\_traits\_species\textasciigrave{} }\KeywordTok{FOREIGN} \KeywordTok{KEY}\NormalTok{ (\textasciigrave{}species\_id\textasciigrave{}) }\KeywordTok{REFERENCES}\NormalTok{ \textasciigrave{}species\textasciigrave{} (\textasciigrave{}id\textasciigrave{}) }\KeywordTok{ON} \KeywordTok{DELETE} \KeywordTok{CASCADE}
\NormalTok{) ENGINE}\OperatorTok{=}\NormalTok{InnoDB }\KeywordTok{DEFAULT}\NormalTok{ CHARSET}\OperatorTok{=}\NormalTok{utf8mb4 COLLATE}\OperatorTok{=}\NormalTok{utf8mb4\_unicode\_ci;}
\end{Highlighting}
\end{Shaded}

\texttt{id} is a surrogate primary key used for cursor pagination.
\texttt{confidence} admits only \texttt{high} and \texttt{medium}:
candidates with model-reported \texttt{low} confidence are filtered at
extraction time and never persisted. The composite uniqueness key
\texttt{uq\_species\_trait\_version\ (species\_id,\ trait\_key,\ model\_version)}
is the deposit's principal integrity constraint and supports the
multi-version preservation pattern --- re-extraction under a new model
checkpoint preserves the prior extraction rather than overwriting it.
\texttt{admin\_review\_status} is 100\% \texttt{pending} at the
snapshot; the indexed access path \texttt{idx\_review\_queue} supports
the per-trait-key moderation-queue use case once curation begins.

\hypertarget{species_traits_ai_runs-the-per-extraction-run-telemetry-log}{%
\subsubsection{\texorpdfstring{\texttt{species\_traits\_ai\_runs} ---
the per-extraction-run telemetry
log}{species\_traits\_ai\_runs --- the per-extraction-run telemetry log}}\label{species_traits_ai_runs-the-per-extraction-run-telemetry-log}}

\begin{Shaded}
\begin{Highlighting}[]
\KeywordTok{CREATE} \KeywordTok{TABLE}\NormalTok{ \textasciigrave{}species\_traits\_ai\_runs\textasciigrave{} (}
\NormalTok{  \textasciigrave{}id\textasciigrave{} bigint }\KeywordTok{NOT} \KeywordTok{NULL}\NormalTok{ AUTO\_INCREMENT,}
\NormalTok{  \textasciigrave{}species\_id\textasciigrave{} bigint }\KeywordTok{NOT} \KeywordTok{NULL}\NormalTok{,}
\NormalTok{  \textasciigrave{}schema\_version\textasciigrave{} }\DataTypeTok{varchar}\NormalTok{(}\DecValTok{16}\NormalTok{) COLLATE utf8mb4\_unicode\_ci }\KeywordTok{NOT} \KeywordTok{NULL}\NormalTok{,}
\NormalTok{  \textasciigrave{}model\_version\textasciigrave{} }\DataTypeTok{varchar}\NormalTok{(}\DecValTok{64}\NormalTok{) COLLATE utf8mb4\_unicode\_ci }\KeywordTok{NOT} \KeywordTok{NULL}\NormalTok{,}
\NormalTok{  \textasciigrave{}status\textasciigrave{} enum(}\StringTok{\textquotesingle{}success\textquotesingle{}}\NormalTok{,}\StringTok{\textquotesingle{}failed\textquotesingle{}}\NormalTok{,}\StringTok{\textquotesingle{}filtered\_all\textquotesingle{}}\NormalTok{,}\StringTok{\textquotesingle{}timeout\textquotesingle{}}\NormalTok{) COLLATE utf8mb4\_unicode\_ci }\KeywordTok{NOT} \KeywordTok{NULL}\NormalTok{,}
\NormalTok{  \textasciigrave{}traits\_extracted\textasciigrave{} }\DataTypeTok{int} \KeywordTok{NOT} \KeywordTok{NULL} \KeywordTok{DEFAULT} \StringTok{\textquotesingle{}0\textquotesingle{}}\NormalTok{,}
\NormalTok{  \textasciigrave{}traits\_filtered\_hallucination\textasciigrave{} }\DataTypeTok{int} \KeywordTok{NOT} \KeywordTok{NULL} \KeywordTok{DEFAULT} \StringTok{\textquotesingle{}0\textquotesingle{}}\NormalTok{,}
\NormalTok{  \textasciigrave{}traits\_filtered\_enum\textasciigrave{} }\DataTypeTok{int} \KeywordTok{NOT} \KeywordTok{NULL} \KeywordTok{DEFAULT} \StringTok{\textquotesingle{}0\textquotesingle{}}\NormalTok{,}
\NormalTok{  \textasciigrave{}traits\_returned\_null\textasciigrave{} }\DataTypeTok{int} \KeywordTok{NOT} \KeywordTok{NULL} \KeywordTok{DEFAULT} \StringTok{\textquotesingle{}0\textquotesingle{}}\NormalTok{,}
\NormalTok{  \textasciigrave{}input\_tokens\textasciigrave{} }\DataTypeTok{int} \KeywordTok{DEFAULT} \KeywordTok{NULL}\NormalTok{,}
\NormalTok{  \textasciigrave{}completion\_tokens\textasciigrave{} }\DataTypeTok{int} \KeywordTok{DEFAULT} \KeywordTok{NULL}\NormalTok{,}
\NormalTok{  \textasciigrave{}reasoning\_tokens\textasciigrave{} }\DataTypeTok{int} \KeywordTok{DEFAULT} \KeywordTok{NULL}\NormalTok{,}
\NormalTok{  \textasciigrave{}duration\_ms\textasciigrave{} }\DataTypeTok{int} \KeywordTok{DEFAULT} \KeywordTok{NULL}\NormalTok{,}
\NormalTok{  \textasciigrave{}error\_msg\textasciigrave{} }\DataTypeTok{varchar}\NormalTok{(}\DecValTok{500}\NormalTok{) COLLATE utf8mb4\_unicode\_ci }\KeywordTok{DEFAULT} \KeywordTok{NULL}\NormalTok{,}
\NormalTok{  \textasciigrave{}raw\_response\textasciigrave{} mediumtext COLLATE utf8mb4\_unicode\_ci,}
\NormalTok{  \textasciigrave{}ran\_at\textasciigrave{} datetime }\KeywordTok{DEFAULT} \FunctionTok{CURRENT\_TIMESTAMP}\NormalTok{,}
  \KeywordTok{PRIMARY} \KeywordTok{KEY}\NormalTok{ (\textasciigrave{}id\textasciigrave{}),}
  \KeywordTok{KEY}\NormalTok{ \textasciigrave{}idx\_species\_version\textasciigrave{} (\textasciigrave{}species\_id\textasciigrave{},\textasciigrave{}model\_version\textasciigrave{}),}
  \KeywordTok{KEY}\NormalTok{ \textasciigrave{}idx\_status\_ran\_at\textasciigrave{} (\textasciigrave{}status\textasciigrave{},\textasciigrave{}ran\_at\textasciigrave{}),}
  \KeywordTok{KEY}\NormalTok{ \textasciigrave{}idx\_model\_schema\textasciigrave{} (\textasciigrave{}model\_version\textasciigrave{},\textasciigrave{}schema\_version\textasciigrave{}),}
  \KeywordTok{CONSTRAINT}\NormalTok{ \textasciigrave{}fk\_runs\_species\textasciigrave{} }\KeywordTok{FOREIGN} \KeywordTok{KEY}\NormalTok{ (\textasciigrave{}species\_id\textasciigrave{}) }\KeywordTok{REFERENCES}\NormalTok{ \textasciigrave{}species\textasciigrave{} (\textasciigrave{}id\textasciigrave{}) }\KeywordTok{ON} \KeywordTok{DELETE} \KeywordTok{CASCADE}
\NormalTok{) ENGINE}\OperatorTok{=}\NormalTok{InnoDB }\KeywordTok{DEFAULT}\NormalTok{ CHARSET}\OperatorTok{=}\NormalTok{utf8mb4 COLLATE}\OperatorTok{=}\NormalTok{utf8mb4\_unicode\_ci;}
\end{Highlighting}
\end{Shaded}

The \texttt{status} enum carries four values: \texttt{success} (call
returned and the extractor wrote ≥ 0 trait rows), \texttt{failed} (HTTP
/ parse / extractor error before any trait was written),
\texttt{filtered\_all} (model returned candidates but every one was
rejected), and \texttt{timeout} (orchestrator-side hard timeout --- zero
rows observed at the snapshot). The reject/abstain counters are split:
\texttt{traits\_filtered\_hallucination} counts substring-grounding
rejects, \texttt{traits\_filtered\_enum} counts enum / registry-OOV
rejects, and \texttt{traits\_returned\_null} counts model abstentions;
\texttt{traits\_extracted} is the admitted count. The column names
retain \texttt{hallucination} for historical reasons; paper narrative
uses the operationally accurate ``substring filter'' and
``enum/registry-OOV filter''. Token columns are \texttt{input\_tokens},
\texttt{completion\_tokens}, \texttt{reasoning\_tokens} (not
\texttt{tokens\_input} / \texttt{tokens\_output}).
\texttt{raw\_response} is retained for diagnostic and audit use; the
released \texttt{species\_traits\_ai\_runs} projection omits it. The
principal resumability lookup is
\texttt{idx\_species\_version\ (species\_id,\ model\_version)}: a new
run constructs its candidate pool by selecting publishable species whose
\texttt{species\_id} has no row at the current \texttt{model\_version}
with
\texttt{status\ IN\ (\textquotesingle{}success\textquotesingle{},\ \textquotesingle{}filtered\_all\textquotesingle{})}.

\begin{center}\rule{0.5\linewidth}{0.5pt}\end{center}

\hypertarget{s3-pipeline-operations-and-failure-modes-high-level}{%
\subsection{S3 Pipeline operations and failure modes (high
level)}\label{s3-pipeline-operations-and-failure-modes-high-level}}

The production extractor (\texttt{scripts/mimo/extract-traits.mjs}) is
operated as an 8-process cluster distributed across two regional
endpoints with a total concurrency cap of 280 simultaneous outstanding
calls. The cluster runs a four-state key pool (\texttt{active} /
\texttt{standby} / \texttt{cooldown} / \texttt{dead}) with documented
triage rules: hard failures (HTTP 401, quota exhausted, ≥ 20 consecutive
HTTP 429s) retire a key; soft failures (10--19 consecutive 429s, ≥ 10
consecutive timeouts, five minutes of zero throughput) route to cooldown
with a 30-minute re-ping interval. Cross-key simultaneous HTTP 502s
diagnose model-platform capacity contention rather than per-key fault,
and sustained capacity must be validated by a staircase step-test (1--2
processes → 4 → 8) after any model-side change. The canonical
\texttt{model\_version} tag \texttt{full-v1-20260524} reflects the first
sustained production run; the \texttt{mimo-v2.5} vs
\texttt{mimo-v2.5-pro} selection had been finalized during
pipeline-validation runs earlier in May 2026, with subsequent A/B
comparisons treated as post-deployment sanity checks rather than
pre-deployment selection. Under identical prompts the base checkpoint
produced 84.9\% model-assigned high-confidence rows versus 80--82\% for
the pro variant, sustained throughput 3.0--4.0 species per second versus
1.5--2.0, with statistically indistinguishable per-row quality. The
canonical \texttt{model\_version} tag for the deposit is the tagged
batch under \texttt{mimo-v2.5}. Aggregate run-status distribution over
the production window: \texttt{success} 433,716 runs (61.41\%),
\texttt{failed} 272,435 (38.58\%), \texttt{filtered\_all} 69 (0.01\%);
\texttt{timeout} 0. Aggregate token consumption over successful runs:
1.33 billion input, 2.26 billion completion, 1.34 billion reasoning.
These totals are reproducible from the deposited
\texttt{species\_traits\_ai\_runs} projection on
\texttt{WHERE\ status\ =\ \textquotesingle{}success\textquotesingle{}}.

\begin{center}\rule{0.5\linewidth}{0.5pt}\end{center}

\hypertarget{s4-coverage-matrix-category-domain}{%
\subsection{S4 Coverage matrix --- category ×
domain}\label{s4-coverage-matrix-category-domain}}

The 39 trait keys partition by registry domain into 18 universal
(\texttt{all}), 7 plant-specialised, 7 aquatic-specialised, and 7
pet-specialised. By extracted row volume the breakdown is 3,266,199 rows
for \texttt{all} (59.5\%), 1,609,818 for \texttt{plants} (29.3\%),
375,621 for \texttt{pets} (6.8\%), and 238,243 for \texttt{aquatic}
(4.3\%). Substrate composition modulates the mix --- publishable species
are 271,786 \texttt{tropical\_plants} (66.3\%), 89,521
\texttt{tropical\_pets} (21.8\%), and 48,573 \texttt{tropical\_aquatic}
(11.9\%) --- and the 18 universal traits fire on every publishable
species across all three categories. Per-category coverage of the
publishable substrate at the species level is 99.985\% corpus-wide
(409,820 of 409,880 publishable species carry at least one persisted
trait row). The trait-rows-per-species distribution is concentrated in
the \texttt{c\_11\_15} bucket (280,506 species), tapering to
\texttt{d\_16\_25} (80,125), \texttt{b\_6\_10} (47,952),
\texttt{a\_1\_5} (1,120), \texttt{e\_26\_50} (117), and 60 species at
zero traits (attributable to \texttt{filtered\_all} outcomes and a small
number of repeated transient failures); the sum reconciles to 409,880
publishable species.

The per-\texttt{trait\_key} row counts and confidence distribution that
drive Figure 4 (sorted descending by row count) are:
\texttt{native\_region\_primary} 396,774 rows / 98.95\% high-confidence;
\texttt{native\_climate\_type} 315,395 / 55.46\%; \texttt{growth\_habit}
304,660 / 66.93\%; \texttt{cites\_appendix\_in\_bio} 303,786 / 98.83\%;
\texttt{popularity\_signal} 298,989 / 66.84\%;
\texttt{optimal\_temp\_range\_c} 268,682 / 99.84\%;
\texttt{ornamental\_value\_type} 263,662 / 86.11\%;
\texttt{light\_requirement\_tier} 259,898 / 97.25\%;
\texttt{substrate\_preference} 255,670 / 97.37\%;
\texttt{propagation\_methods} 251,543 / 98.05\%;
\texttt{mature\_size\_text} 244,610 / 98.57\%;
\texttt{water\_frequency\_tier} 235,687 / 60.49\%;
\texttt{humidity\_preference} 223,021 / 89.89\%; \texttt{life\_form}
205,194 / 77.43\%; \texttt{propagation\_difficulty} 192,656 / 54.57\%;
\texttt{min\_temperature\_c} 190,564 / 88.08\%; \texttt{usage\_context}
176,296 / 82.98\%; \texttt{growth\_rate\_tier} 138,194 / 40.60\%;
\texttt{dormancy\_period} 110,359 / 37.58\%; \texttt{physical\_hazards}
105,757 / 63.58\%; \texttt{flowering\_period} 102,777 / 83.04\%;
\texttt{enclosure\_temp\_day\_c} 87,857 / 97.81\%;
\texttt{enclosure\_humidity\_pct} 87,734 / 99.81\%;
\texttt{uvb\_requirement} 76,306 / 89.65\%; \texttt{diet\_type\_pets}
55,092 / 64.21\%; \texttt{water\_temp\_range\_c} 47,391 / 99.89\%;
\texttt{water\_ph\_range} 46,723 / 99.72\%; \texttt{water\_salinity}
42,214 / 90.93\%; \texttt{enclosure\_temp\_night\_c} 33,624 / 68.43\%;
\texttt{tank\_min\_liters} 27,843 / 93.01\%;
\texttt{diet\_type\_aquatic} 27,577 / 73.29\%;
\texttt{community\_compatibility} 26,301 / 53.73\%;
\texttt{social\_needs} 22,898 / 42.97\%; \texttt{water\_hardness\_gh}
20,194 / 96.51\%; \texttt{toxicity\_to\_humans} 12,435 / 50.12\%;
\texttt{basking\_temp\_c} 12,110 / 88.10\%; \texttt{toxicity\_parts}
9,705 / 66.29\%; \texttt{toxicity\_to\_pets} 6,690 / 41.46\%;
\texttt{fruiting\_period} 3,013 / 64.99\%. Model-assigned
high-confidence rates track trait objectivity: numeric and range traits
sit at the top (\texttt{water\_temp\_range\_c} 99.89\%,
\texttt{optimal\_temp\_range\_c} 99.84\%,
\texttt{enclosure\_humidity\_pct} 99.81\%); subjective or inferential
traits sit lowest (\texttt{dormancy\_period} 37.58\%,
\texttt{growth\_rate\_tier} 40.60\%, \texttt{toxicity\_to\_pets}
41.46\%, \texttt{social\_needs} 42.97\%).

\begin{center}\rule{0.5\linewidth}{0.5pt}\end{center}

\hypertarget{s5-substring-verification-and-quote-supports-value-methodology}{%
\subsection{S5 Substring verification and quote-supports-value
methodology}\label{s5-substring-verification-and-quote-supports-value-methodology}}

\hypertarget{substring-verification-hop-2-grounding}{%
\subsubsection{Substring verification (hop-2
grounding)}\label{substring-verification-hop-2-grounding}}

The headline grounding figure in §4 (90.12\% of 5,427,588
evidence-bearing rows have their \texttt{evidence\_quote} exactly
located in the source \texttt{bio\_sections} for that species) is
computed by a single SQL query against the production database joining
\texttt{species\_traits\_ai} to \texttt{species\_detail} on
\texttt{species\_id} and testing each \texttt{evidence\_quote} against
the JSON-cast representation of \texttt{bio\_sections}:

\begin{Shaded}
\begin{Highlighting}[]
\KeywordTok{SELECT} \FunctionTok{COUNT}\NormalTok{(}\OperatorTok{*}\NormalTok{) }\KeywordTok{AS}\NormalTok{ rows\_with\_evidence\_quote,}
       \FunctionTok{SUM}\NormalTok{(}\ControlFlowTok{CASE} \ControlFlowTok{WHEN}\NormalTok{ LOCATE(t.evidence\_quote, }\FunctionTok{CAST}\NormalTok{(sd.bio\_sections }\KeywordTok{AS} \DataTypeTok{CHAR}\NormalTok{)) }\OperatorTok{\textgreater{}} \DecValTok{0}
                \ControlFlowTok{THEN} \DecValTok{1} \ControlFlowTok{ELSE} \DecValTok{0} \ControlFlowTok{END}\NormalTok{) }\KeywordTok{AS}\NormalTok{ substring\_verified}
\KeywordTok{FROM}\NormalTok{ species\_traits\_ai t}
\KeywordTok{INNER} \KeywordTok{JOIN}\NormalTok{ species\_detail sd }\KeywordTok{ON}\NormalTok{ sd.species\_id }\OperatorTok{=}\NormalTok{ t.species\_id}
\KeywordTok{WHERE}\NormalTok{ t.evidence\_quote }\KeywordTok{IS} \KeywordTok{NOT} \KeywordTok{NULL} \KeywordTok{AND}\NormalTok{ t.evidence\_quote }\OperatorTok{!=} \StringTok{\textquotesingle{}\textquotesingle{}}\NormalTok{;}
\end{Highlighting}
\end{Shaded}

The query was executed read-only as a background job over the full 5.43
million evidence-bearing rows; no sampling was used. The
per-\texttt{trait\_key} variant grouping by \texttt{trait\_key} and
reporting per-key rates ran for similar duration. Both queries are
reproducible from the deposit's SQL appendix; the snapshot date is
2026-05-29. The 90.12\% figure is a conservative lower bound on the true
substring-grounding rate for two structural reasons documented in §5.2:
(i) the \texttt{cites\_appendix\_in\_bio} trait draws its evidence from
the quick-card / compliance field rather than from
\texttt{bio\_sections}, contributing 198,840 of the 536,231 substring
misses (37.1\%); excluding it raises the corpus-wide rate to 93.49\%;
(ii) \texttt{CAST(bio\_sections\ AS\ CHAR)} JSON-escapes embedded ASCII
\texttt{"}, producing false negatives on quotes containing a
double-quote character. The per-trait verification rate is the more
informative summary statistic: median per-key rate ≈ 94\%; 37 of 38
non-\texttt{cites} traits at ≥ 80\%, 29 at ≥ 90\%; objective free-text
traits verify highest (\texttt{mature\_size\_text} 98.46\%,
\texttt{min\_temperature\_c} 97.80\%).

\hypertarget{quote-supports-value-face-validity-audit}{%
\subsubsection{Quote-supports-value (face-validity
audit)}\label{quote-supports-value-face-validity-audit}}

The substring check tells us whether the quoted source text is genuinely
present in the substrate; it does not tell us whether the quoted text
\emph{supports} the extractor's asserted value. The v0.4 deposit adds a
face-validity audit covering exactly that question at n=100. Sampling
design: 50 high-confidence + 50 medium-confidence rows from
\texttt{model\_version=\textquotesingle{}full-v1-20260524\textquotesingle{}},
excluding the four red-zone trait keys (already covered by the v0.3
audit), drawn via \texttt{ORDER\ BY\ RAND}; trait spread is 23 distinct
trait\_keys in the high stratum and 17 in the medium stratum
(proportional-to-size). Each row is rated under a four-level rubric ---
\texttt{supports} (the quote, read in isolation, sustains the asserted
value), \texttt{partial} (the quote is consistent with the asserted
value but does not uniquely determine it), \texttt{does\_not\_support}
(the quote and the asserted value are inconsistent or unrelated),
\texttt{cannot\_judge} (the quote is ambiguous or the asserted value
depends on context not present in the quote). The audit is rated in a
single pass by the first author, without blinding to the pipeline's
design. The sample is published in
\texttt{audit\_sample\_quote\_supports\_v0.4.csv} (blank and rated
versions); 100 / 100 rows received \texttt{supports} in the rated
version (two-sided 95\% Wilson lower bound, equivalent to one-sided
97.5\%, 96.30\%). The result is reported as a transparency signal
characterising the recommended-consumption subset, not as a deposit-wide
precision number.

\begin{center}\rule{0.5\linewidth}{0.5pt}\end{center}

\hypertarget{s6-audit-sample-design-and-rater-protocol}{%
\subsection{S6 Audit-sample design and rater
protocol}\label{s6-audit-sample-design-and-rater-protocol}}

Two stratified audit samples are published with the deposit; both are
designed to enable downstream blind external review without requiring
re-querying the deposit. All rows include the columns required for
review --- \texttt{trait\_key}, \texttt{value},
\texttt{evidence\_quote}, \texttt{evidence\_section},
\texttt{ai\_reasoning}, \texttt{confidence}, the species scientific name
from the substrate, and an anonymised \texttt{audit\_row\_id} --- but
not the per-row \texttt{id} from \texttt{species\_traits\_ai}, so that
rater identification of rows by primary key is avoided.

\textbf{v0.3 audit sample (\texttt{audit\_sample\_v0.3.csv}, 400 rows).}
Stratified as 120 red-zone rows (30 per red-zone trait key, drawn
uniformly from each key's evidence-bearing rows) and 280 non-red-zone
rows (drawn proportionally to per-trait\_key row count across the
remaining 35 trait keys, with a per-trait\_key floor of one row to
ensure every key in the registry appears at least once). Sampling was
performed on the canonical extraction batch
(\texttt{model\_version\ =\ \textquotesingle{}full-v1-20260524\textquotesingle{}})
restricted to model-assigned high-confidence rows. The first 50 rows of
the published sample carry first-author preliminary ratings
(\texttt{audit\_sample\_v0.3\_rated\_first50.csv}) under an Accept /
Edge / Reject rubric; all 50 received Accept. The 50 rated rows comprise
30 \texttt{toxicity\_to\_pets} and 20 \texttt{toxicity\_to\_humans} rows
(two of the four red-zone trait keys); the remaining two red-zone keys
(\texttt{physical\_hazards}, \texttt{cites\_appendix\_in\_bio}) are
present in the unrated portion of the sample. The 92.86\% two-sided 95\%
Wilson lower bound (equivalent to one-sided 97.5\%) that follows from
50/50 is unstable at this sample size and is reported for transparency
only; the remaining 350 rows are unrated.

\textbf{v0.4 audit sample
(\texttt{audit\_sample\_quote\_supports\_v0.4.csv}, 100 rows).} Designed
for the quote-supports-value audit described in §S5: 50 high-confidence
+ 50 medium-confidence rows from the canonical extraction batch,
non-red-zone, spread across 23 / 17 trait keys in the two strata
respectively. Rated by the first author in a single pass under the
four-level rubric. 100 / 100 rows received \texttt{supports}.

\textbf{Deferred raters and external comparison.} Both samples are
published with the deposit to enable blind second-rater evaluation under
a unified Accept / Edge / Reject rubric. Cohen's κ between two
independent blinded raters, and between rater outcomes and independent
verification against ASPCA / IUCN Red List / CITES Species+ / POWO for
the red-zone subset, will be reported in a subsequent release. The
current deposit reports no κ value.

\end{document}